%% file: hmoe.tex
\documentclass[10pt,twocolumn,letterpaper]{article}

\usepackage[pagenumbers]{cvpr} 

\input{preamble}

\usepackage{bm}
\usepackage{stfloats}
\usepackage{float}
\usepackage{subcaption}
\usepackage{graphicx}
\usepackage{amsmath}
\usepackage{amssymb}
\newcommand{\meddot}{\scalebox{0.6}{$\bullet$}}

\usepackage{ulem}
\usepackage{multirow}
\usepackage{booktabs}
\usepackage{colortbl}
\definecolor{middlegrey}{rgb}{0.375,0.375,0.375}

\usepackage{arydshln}
\usepackage{wrapfig}
\usepackage[T1]{fontenc}

\usepackage{algorithm}
\usepackage{algpseudocode}
\algnewcommand\algorithmicforeach{\textbf{for each}}
\algdef{S}[FOR]{ForEach}[1]{\algorithmicforeach\ #1\ \algorithmicdo}

\definecolor{cvprblue}{rgb}{0.21,0.49,0.74}
\usepackage[pagebackref,breaklinks,colorlinks,citecolor=cvprblue]{hyperref}

\def\confName{CVPR}
\def\confYear{2024}

\title{HMOE: Hypernetwork-based Mixture of Experts for Domain Generalization}

\author{
		Jingang Qu\textsuperscript{1} \quad
		Thibault Faney\textsuperscript{2} \quad
		Ze Wang\textsuperscript{1} \quad
		Patrick Gallinari\textsuperscript{1,3} \\
		Soleiman Yousef\textsuperscript{2} \quad
		Jean-Charles de Hemptinne\textsuperscript{2}
		\\[.5em]
		Sorbonne Université, CNRS, ISIR, 75005 Paris, France\textsuperscript{1} \quad
		IFPEN \textsuperscript{2} \quad 
		Criteo AI Lab, Paris, France \textsuperscript{3} 
		\\
}

\begin{document}
	\maketitle

\begin{abstract}
Due to domain shifts, machine learning systems typically struggle to generalize well to new domains that differ from those of training data, which is what domain generalization (DG) aims to address. Although a variety of DG methods have been proposed, most of them fall short in interpretability and require domain labels, which are not available in many real-world scenarios. This paper presents a novel DG method, called HMOE: Hypernetwork-based Mixture of Experts (MoE), which does not rely on domain labels and is more interpretable. MoE proves effective in identifying heterogeneous patterns in data. For the DG problem, heterogeneity arises exactly from domain shifts. HMOE employs hypernetworks taking vectors as input to generate the weights of experts, which promotes knowledge sharing among experts and enables the exploration of their similarities in a low-dimensional vector space. We benchmark HMOE against other DG methods under a fair evaluation framework -- DomainBed. Our extensive experiments show that HMOE can effectively separate mixed-domain data into distinct clusters that are surprisingly more consistent with human intuition than original domain labels. Using self-learned domain information, HMOE achieves state-of-the-art results on most datasets and significantly surpasses other DG methods in average accuracy across all datasets.
\end{abstract}

\section{Introduction}
\label{sec:intro}

Domain generalization (DG) aims to train models on known domains to perform well on unseen domains, which is crucial for deploying models in safety-critical applications. Over the past decade, a variety of DG algorithms have been proposed \cite{gulrajaniSearchLostDomain2020, zhouDomainGeneralizationSurvey2022, wangGeneralizingUnseenDomains2022}, focusing primarily on developing DG-specific data augmentation techniques and learning domain-invariant representations to build generalizable predictors. However, many high-performing DG algorithms rely on domain labels to explicitly reduce inter-domain differences, severely limiting their applicability in real-world scenarios where domain annotation may be prohibitively expensive. Additionally, current DG algorithms lack interpretability and cannot provide insight into the causes of success or failure in generalizing to new domains. Therefore, this work aims to develop a novel DG algorithm that does not require domain labels and is more interpretable.

We follow the nomenclature established by \cite{chenCompoundDomainGeneralization2022}, which refers to DG with domain labels as vanilla DG and the more challenging DG without domain labels as compound DG. This work focuses on addressing compound DG by inferring latent domains from mixed-domain data and using them effectively. \cite{deshmukhGeneralizationErrorBound2019, blanchardDomainGeneralizationMarginal2021, muandetDomainGeneralizationInvariant2013} demonstrated that using domain-wise datasets can theoretically yield lower generalization error bounds and better DG performance compared to using mixed data directly, indicating the importance of domain information. Furthermore, latent domain discovery helps us understand the workings of models and enhances interpretability. To make the problem tractable, we assume that latent domains are distinct and separable.

In this paper, we introduce HMOE: \textbf{H}ypernetwork-based Mixture of Experts (\textbf{MoE}). MoE is a well-established learning paradigm that aggregates a number of experts by calculating the weighted sum of their predictions \cite{jacobsAdaptiveMixturesLocal1991, jordanHierarchicalMixturesExperts1994}, where the aggregation weights, commonly referred to as gate values, are determined by a routing mechanism and add up to 1. HMOE capitalizes on MoE's \emph{divide and conquer} property, that is, the routing mechanism can softly partition the input space into subspaces in an unsupervised manner during training \cite{yukselTwentyYearsMixture2012}, with each subspace assigned to an expert. We further expect that each subspace is associated with a latent domain, enabling latent domain discovery. During inference, we can compare the similarities between an unseen test domain and the inferred domains based on gate values, hence improving interpretability. \cite{guoMultisourceDomainAdaptation2018, zhongMetaDMoEAdaptingDomain2022} have validated MoE in domain adaptation \cite{wangDeepVisualDomain2018} and showed that MoE can leverage the specialty of individual domain and alleviate negative knowledge transfer \cite{standleyWhichTasksShould2020} compared to using a single model to learn different domains concurrently.

HMOE innovatively uses a neural network, called hypernetwork \cite{haHypernetworks2016}, which takes vectors as input to generate the weights for MoE's experts. By mapping vectors to experts, hypernetworks enable the exploration of experts' similarities in a low-dimensional vector space, facilitating latent domain discovery. Hypernetworks also serve as a bridge between experts and provide them a channel to exchange information, thereby promoting knowledge sharing.

MoE's intrinsic soft partitioning is not always effective and sometimes fails to maintain a consistent data division, especially when the distinction between latent domains is not significant. To address this issue, we propose a differentiable dense-to-sparse Top-1 routing algorithm, which forces gate values to become one-hot and converges to hard partitioning. This leads to sparse-gated MoE, which improves and stabilizes latent domain discovery. In addition, to better incorporate hypernetworks into MoE, we introduce an embedding space that contains a set of learnable embedding vectors corresponding one-to-one with experts. This embedding space is fed to hypernetworks to generate the weights of experts and is also part of the routing mechanism to compute gate values, thus enhancing the interaction between hypernetworks and the routing mechanism.

We also propose an intra-domain \emph{mixup} to further improve the generalization ability of HMOE. \emph{mixup} creates virtual training samples by taking a linear combination of two randomly chosen inputs and their labels \cite{zhangMixupEmpiricalRisk2017}, and we perform \emph{mixup} within each inferred latent domain.

Our contributions are as follows:
(1) We present a novel DG method -- HMOE within the framework of MoE, that does not require domain labels, enables latent domain discovery, and offers excellent interpretability.
(2) HMOE leverages hypernetworks to generate expert weights and achieves sparse-gated MoE.
(3) As far as we know, HMOE is the first work that can jointly learn and use latent domains in an end-to-end way.
(4) Extensive experiments are conducted to compare HMOE with other DG methods under a fair evaluation framework -- DomainBed \cite{gulrajaniSearchLostDomain2020}. HMOE exhibits state-of-the-art performance on most datasets and greatly outperforms other DG methods in average accuracy.

\section{Related Work}
\label{sec:related}

\subsection{Domain Generalization (DG)}
\label{sec:related_DG}
The goal of DG is to train a predictor on known domains that can generalize well to unseen domains.

\noindent \textbf{Vanilla DG} \quad The first line of work is to design DG-specific data augmentation techniques to increase the diversity and quantity of training data to improve DG performance \cite{yueDomainRandomizationPyramid2019, volpiGeneralizingUnseenDomains2018, shankarGeneralizingDomainsCrossgradient2018, zhangMixupEmpiricalRisk2017, liuUnifiedFeatureDisentangler2018, zhouDomainGeneralizationMixstyle2021, qiaoLearningLearnSingle2020, zhouLearningGenerateNovel2020}. Previous work learned domain-invariant representations through invariant risk minimization \cite{arjovskyInvariantRiskMinimization2019, kruegerOutofdistributionGeneralizationRisk2021, ahujaInvariancePrincipleMeets2021}, kernel methods \cite{muandetDomainGeneralizationInvariant2013, ghifaryScatterComponentAnalysis2016, ganLearningAttributesEquals2016, blanchardDomainGeneralizationMarginal2021}, feature alignment \cite{panDomainAdaptationTransfer2010, tzengDeepDomainConfusion2014, wangVisualDomainAdaptation2018, sunDeepCoralCorrelation2016, pengMomentMatchingMultisource2019, liDomainGeneralizationAdversarial2018, motiianUnifiedDeepSupervised2017, ghifaryDomainGeneralizationObject2015, matsuuraDomainGeneralizationUsing2020}, and domain-adversarial training \cite{ganinUnsupervisedDomainAdaptation2015, ganinDomainadversarialTrainingNeural2016, liDomainGeneralizationAdversarial2018, liDeepDomainGeneralization2018, gongDlowDomainFlow2019}. Another approach is to disentangle latent features into class-specific and domain-specific representations \cite{khoslaUndoingDamageDataset2012, pengDomainAgnosticLearning2019, ilseDivaDomainInvariant2020, namReducingDomainGap2021, zhangPrincipledDisentanglementDomain2022}. General machine learning paradigms were also applied to vanilla DG, such as meta-learning \cite{liLearningGeneralizeMetalearning2018, balajiMetaregDomainGeneralization2018, douDomainGeneralizationModelagnostic2019, liEpisodicTrainingDomain2019}, self-supervised learning \cite{carlucciDomainGeneralizationSolving2019, kimSelfregSelfsupervisedContrastive2021}, gradient manipulation \cite{huangSelfchallengingImprovesCrossdomain2020, shiGradientMatchingDomain2021, rameFishrInvariantGradient2022}, and distributionally robust optimization \cite{sagawaDistributionallyRobustNeural2020, kruegerOutofdistributionGeneralizationRisk2021}.

\noindent \textbf{Compound DG} \quad There are some DG algorithms that do not require domain labels by design \cite{huangSelfchallengingImprovesCrossdomain2020, matsuuraDomainGeneralizationUsing2020, liSimpleFeatureAugmentation2021, namReducingDomainGap2021, zhangPrincipledDisentanglementDomain2022, chenCompoundDomainGeneralization2022}. Besides improving DG performance, latent domain discovery is also an important task for compound DG and contributes to better interpretability. \cite{matsuuraDomainGeneralizationUsing2020, chenCompoundDomainGeneralization2022} can do this but have two main limitations: (1) Their methods proceed in two phases: first infer latent domains from mixed data and then deal with DG using the inferred domains, which is similar to vanilla DG. The problem is that the second phase depends on the first and cannot provide some feedback to correct possible errors in domain discovery. (2) Their methods assume that domain shift arises from stylistic differences to identify latent domains, which does not always hold. 

On the contrary, HMOE is trained in an end-to-end manner and leverages MoE to discover latent domains without an explicit induced bias on the cause of domain shift.

\subsection{Hypernetworks}
\label{sec:related_hypernet}

A hypernetwork is a neural network that generates the weights of another target network. Hypernetworks were initially proposed by \cite{haHypernetworks2016} and have since been applied to optimization problems \cite{lorraineStochasticHyperparameterOptimization2018, navonLearningParetoFront2020}, meta-learning \cite{zhaoMetalearningHypernetworks2020}, continuous learning \cite{vonoswaldContinualLearningHypernetworks2019, brahmaHypernetworksContinualSemiSupervised2021}, multi-task learning \cite{linControllableParetoMultitask2020, tayHypergridTransformersSingle2021, mahabadiParameterefficientMultitaskFinetuning2021}, few-shot learning \cite{senderaHypershotFewshotLearning2022}, and federated learning \cite{shamsianPersonalizedFederatedLearning2021}.

\subsection{Mixture of Experts (MoE)}
\label{sec:related_moe}

\begin{figure}[htbp]
	\centering
	\begin{subfigure}{0.57\linewidth}
		\includegraphics[width=\textwidth]{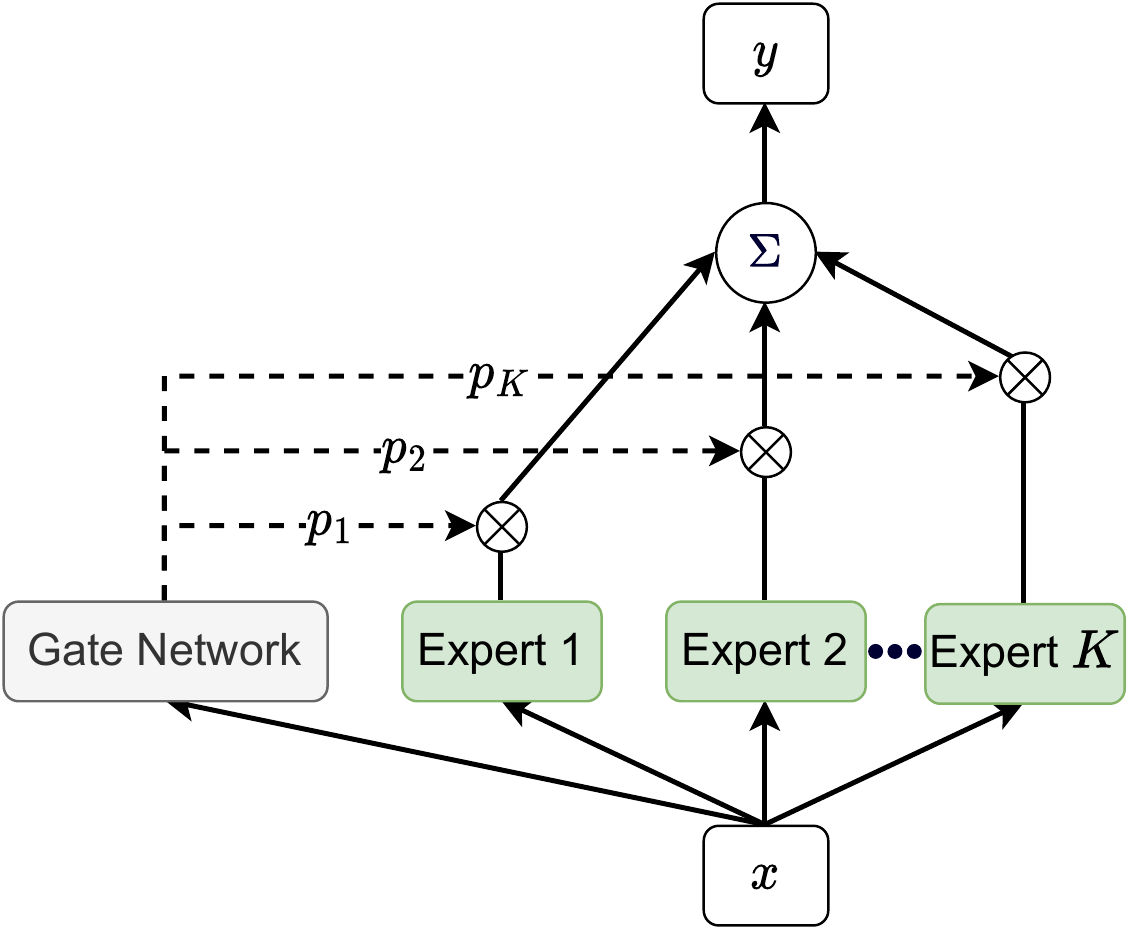}
		\caption{Classical MoE}
		\label{fig:moe}
	\end{subfigure}
	\begin{subfigure}{0.42\linewidth}
		\includegraphics[width=\textwidth]{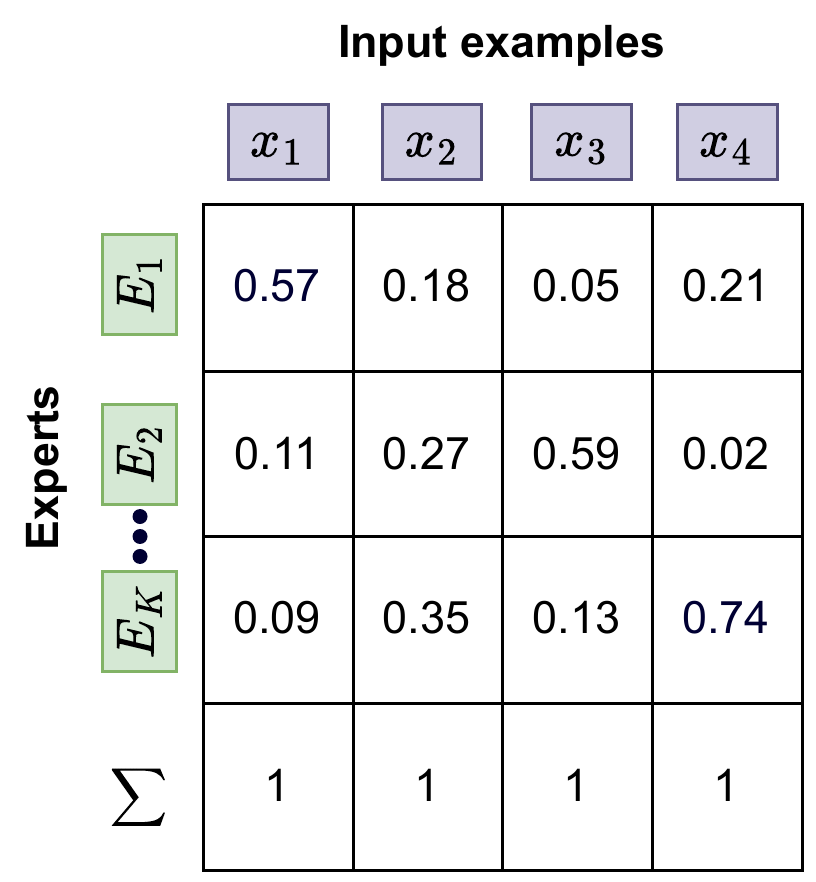}
		\caption{Gate value matrix}
		\label{fig:gate_values}
	\end{subfigure}
	\caption{(a) MoE calculates the weighted sum of experts' outputs. (b) Gate values are determined by a gate network.}
	\label{fig:moe_and_gate}
\end{figure}

MoE was originally proposed by \cite{jacobsAdaptiveMixturesLocal1991, jordanHierarchicalMixturesExperts1994} and consists of two main components: experts and a gate network, as shown in \cref{fig:moe_and_gate}. The output of MoE is the weighted sum of experts, with gate values calculated by the gate network on a per-example basis. In recent years, MoE has regained attention as a way to scale up deep learning models and more efficiently harness modern hardware \cite{shazeerOutrageouslyLargeNeural2017, lepikhinGshardScalingGiant2020, fedusSwitchTransformersScaling2021, duGlamEfficientScaling2022, zophDesigningEffectiveSparse2022, fedusReviewSparseExpert2022}. In this case, sparse MoE is preferred, which routes each example only to the experts with Top-1 or Top-K gate values.

\subsection{Application of Hypernetworks and MoE in DG}

As far as we know, no work has applied hypernetworks to solve DG in computer vision. Recently, \cite{volkExamplebasedHypernetworksOutofdistribution2022} applied hypernetworks to DG in natural language processing (NLP) and achieved SOTA results on two NLP-related DG tasks. 

As for MoE, \cite{liSparseFusionMixtureofExperts2022} proposed replacing feed-forward network layer (FFN) of Vision Transformer (ViT) \cite{dosovitskiyImageWorth16x162020} with a sparse mixture of FFN experts to improve DG performance. \cite{guoMultisourceDomainAdaptation2018, zhongMetaDMoEAdaptingDomain2022} applied MoE to a task similar to DG, namely domain adaptation \cite{wangDeepVisualDomain2018}, but they require domain labels to train an expert for each domain separately. \cite{zhongMetaDMoEAdaptingDomain2022} aggregates the outputs of experts via a transformer-based aggregator, but its aggregator is trained with fixed experts and cannot provide probabilities of experts, while HMOE can do this and is more interpretable. In addition, if we regard MoE as a kind of ensemble method, \cite{manciniBestSourcesForward2018, dinnocenteDomainGeneralizationDomainspecific2018, zhouDomainAdaptiveEnsemble2021} share the same spirit.

\section{Method}
\label{sec:method}

\subsection{Problem Setting}
Let $\mathcal{X}$ denote an input space and $\mathcal{Y}$ a target space. A domain $S$ is characterized by a joint distribution $P^s_{XY}$ on $\mathcal{X} \times \mathcal{Y}$. In vanilla DG setting, we have a training set containing $M$ known domains, \ie, $\mathcal{D}_{tr}^V = \{ \mathcal{D}^s \}_{s=1}^{M}$ with $\mathcal{D}^s = \{(x^s_i, y^s_i, d^s_i)\}_{i=1}^{N_s}$ where $(x^s_i, y^s_i) \sim P^s_{XY}$ and $d^s_i$ is the domain index or label. Also consider a test dataset $\mathcal{D}_{te}$ composed of unknown domains different from those of $\mathcal{D}_{tr}^V$. Vanilla DG aims to train a robust predictor $f: \mathcal{X} \to \mathcal{Y}$ on $\mathcal{D}_{tr}^V$ to achieve a minimum predictive error on $\mathcal{D}_{te}$, \ie, $\min_f \mathbb{E}_{(x, y) \sim \mathcal{D}_{te}} [ \ell(f(x), y) ]$, where $\ell$ is the loss function.

Our work focuses on the more difficult compound DG, for which the training set $\mathcal{D}_{tr} = \{ (x_i, y_i) \}_{i=1}^{N}$ contains mixed domains and has no domain annotation. However, as demonstrated in \cite{gulrajaniSearchLostDomain2020, zhouDomainGeneralizationSurvey2022, wangGeneralizingUnseenDomains2022}, intrinsic inter-domain relationships play a key role in obtaining better generalization performance. Therefore, our proposed HMOE is designed to discover latent domains by dividing $\mathcal{D}_{tr}$ into clusters and to fully leverage the learned domain information in order to perform well on unknown domains.

\subsection{Overall Architecture}
\label{sec:arch}

\begin{figure*}[htbp]
	\centering
	\begin{subfigure}{0.76\linewidth}
		\includegraphics[width=\textwidth]{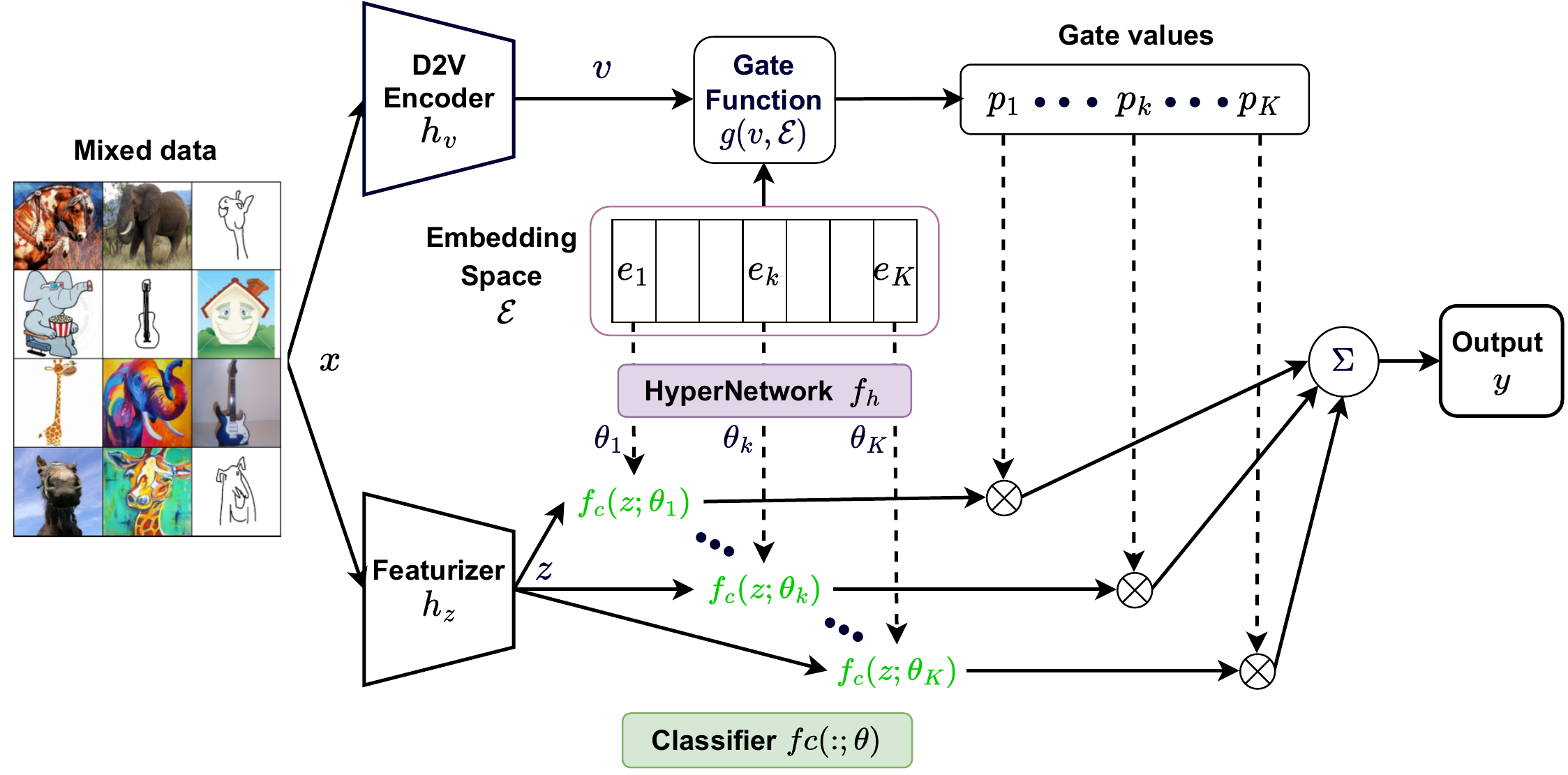}
		\caption{An overview of HMOE}
		\label{fig:arch}
	\end{subfigure}
	\hspace{0.4cm}
	\begin{subfigure}{0.19\linewidth}
		\includegraphics[width=\textwidth]{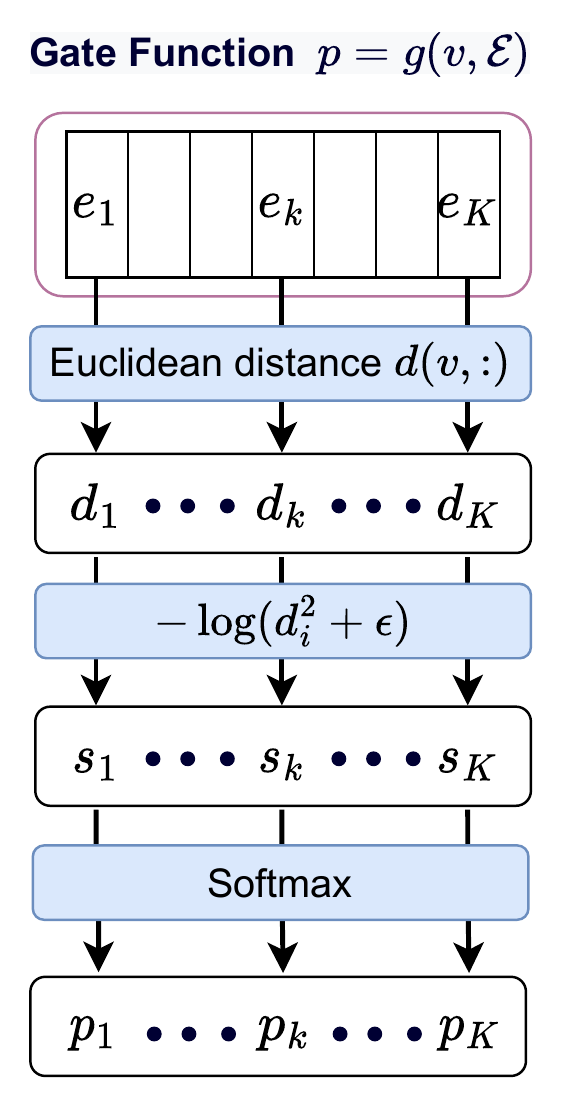}
		\caption{Gate function}
		\label{fig:gate_function}
	\end{subfigure}
	\caption{(a) In the upper branch (\ie, the domain path), the input goes through the D2V encoder into the embedding space, and a predefined gate function calculates gate values. In the lower branch (\ie, the classifier path), the hypernetwork takes the embedding vectors as input to create a set of classifiers. The final output is the weighted sum of the outputs of  classifiers. (b) The gate function determines gate values based on the distances between the output of the D2V encoder and embedding vectors. Smaller distances yield greater gate values.}
	\label{fig:hmoe}
\end{figure*}

An overview of HMOE is illustrated in \cref{fig:arch}. HMOE processes input $x$ through two paths: the domain path for latent domain discovery and the classifier path to train an expert for each latent domain.

The classifier path begins with a featurizer $h_z$ to extract high-level features from $x$, which can be a pretrained network, such as VGG \cite{simonyanVeryDeepConvolutional2014}, ResNet \cite{heDeepResidualLearning2016}, or ViT \cite{dosovitskiyImageWorth16x162020}. We define a discrete learnable embedding space $\mathcal{E}$ consisting of $K$ embedding vectors $\{ e_k \in \mathbb{R}^D \}_{k=1}^K$ ($D$ represents the embedding dimension), each corresponding to a classifier expert. These vectors are fed into a hypernetwork $f_h$ to generate a set of weights $\{ \theta_k \}_{k=1}^K$, which further form a set of experts $\{ f_c(:; \theta_k) \}_{k=1}^K$. The output of the featurizer $z$ is passed to these experts to compute their corresponding outputs, that is, $y_k = f_c(z; \theta_k)$.

The domain path begins with a Domain2Vec (D2V) encoder $h_v$, which transforms $x$ into the embedding space $\mathcal{E}$ and outputs $v \in \mathbb{R}^D$. The output $v$ is then compared with the embedding vectors through a predefined gate function $g(v, \mathcal{E})$, as shown in \cref{fig:gate_function}, to produce a set of probabilities $\bm{p} = \{ p_k \}_{k=1}^K$. The final output of HMOE is the weighted sum of the outputs of experts as follows:
\begin{equation}
    y = \sum_{k=1}^K p_k y_k = g(h_v(x), \mathcal{E}) \ \meddot \left[ f_c(h_z(x); f_h(e_k)) \right]_{k=1}^K \label{eq:output}
\end{equation}

\subsection{Hypernetworks}
\label{sec:hypernet}

We employ a hypernetwork $f_h$ taking a vector $e$ as input to produce weights for classifier $f_c$. In our work, both $f_h$ and $f_c$ are MLPs. Essentially, $f_c$ acts as a computational graph placeholder, $e$ is a conditioning signal, and $f_h$ maps $e$ to a function. The roles of $f_h$ include: (1) easing latent domain discovery, (2) using many experts without a major increase in parameters, (3) offering another interaction between experts and the routing mechanism besides the aggregation of experts compared to the classical MoE, and (4) enabling the generalization of experts beyond aggregation (As we will see later, $f_h$ can directly take the D2V encoder as input).

\subsection{Routing Mechanism}
\label{sec:routing}
 
\subsubsection{Gate Function}
\label{sec:gate_func}

To quantify the responsibilities of experts for each input example and to aggregate experts' outputs, we need to calculate gate values $\bm{p}$. As shown in \cref{fig:gate_function}, based on the output of the D2V encoder $v$ and the embedding space $\mathcal{E}$, we define a gate function $g(v, \mathcal{E})$ to calculate $\bm{p}$ as follows:
\begin{subequations}
    \begin{gather}
        d_k = \lVert v - e_k \rVert _2    \label{eq:dist_ve}   \\
        s_k = -\log (d_k^2 + \epsilon)    \label{eq:neg_log}   \\
        p_k = \frac{\exp (s_k)}{\sum_{j=1}^K \exp (s_j)}
    \end{gather}
\end{subequations}
where $\epsilon$ is a small value. The negative logarithm in \cref{eq:neg_log} is used to establish a negative correlation between $d_k$ and $p_k$ (\ie, the smaller $d_k$, the larger $p_k$) and to nonlinearly rescale the distance $d$ (\ie, stretch small $d$ and squeeze great $d$), which makes $\bm{p}$ less sensitive to large $d$.

\subsubsection{Differentiable Dense-to-Sparse Top-1 Routing}
\label{sec:diff_top1}
Based on gate values $\bm{p}$, the routing mechanism determines where and how to route input examples. A consistent and cohesive routing is crucial to the training stability and convergence of MoE \cite{daiStableMoEStableRouting2022}. In order to stabilize the routing and enhance latent domain discovery to capture less obvious domain differences, sparse-gated MoE is preferable. However, the commonly used Top-1 or Top-K functions are not differentiable and may cause oscillatory behavior of gate values during training \cite{hazimehDselectkDifferentiableSelection2021}. To overcome this limitation, we propose a differentiable dense-to-sparse Top-1 routing algorithm by introducing an entropy loss on $\bm{p}$ as follows:
\begin{equation}
    \mathcal{L}_{en} = \mathbb{E}_{(x, y) \sim \mathcal{D}_{tr}} \left[ \mathbb{H} \big( g(h_v(x), \mathcal{E}) \big) \right]
\end{equation}
where $\mathbb{H(\cdot)}$ denotes the entropy of a distribution. In practice, we multiply $\mathcal{L}_{en}$ by $\gamma_{en}$ that linearly increases from 0 to 1 in the first half of training and remains at 1 in the second. Early on, $\gamma_{en}$ is small, and the distances between $v$ and the embedding vectors are almost the same, leading to a uniform $\bm{p}$. Therefore, all experts can be fully trained and gradually become specialized. In the later stages, $\mathcal{L}_{en}$ forces $\bm{p}$ to become one-hot based on specialized experts.

Due to the negative logarithm in \cref{eq:neg_log}, the D2V encoder has to move towards one embedding vector to minimize $\mathcal{L}_{en}$ instead of moving away from others.

\subsubsection{Expert Load Balancing}
\label{sec:load_balance}
Sparse-gated MoE may suffer from an unbalanced expert load. We define the importance of experts as $I(X) = [I_1(X), \cdots, I_K(X)]$, where $X$ represents a single batch and $I_k(X)$ is specified as the sum of gate values assigned to the $k$th expert (\ie, sum the gate value matrix in \cref{fig:gate_values} along the example dimension). \cite{pavlitskayaBalancingExpertUtilization2022} defines a distribution $P = I(X) / \sum I(X)$ and uses the KL-divergence between $P$ and the uniform distribution $\mathcal{U}$ to balance the expert load, which is also used in our work:
\begin{equation}
	\mathcal{L}_{kl} = D_{KL}(P \Vert \mathcal{U}) = D_{KL} \left( \frac{I(X)}{\sum I(X)} \Vert \mathcal{U} \right)
\end{equation}

\subsection{Embedding Space}
\label{sec:emb_space}
The embedding space $\mathcal{E}$ plays a key role in HMOE. As we can see, the embedding vectors have an effect on both the generation of expert weights and the routing mechanism, thus serving as a bridge to balance these two parts. In addition, these embedding vectors are learnable like the weights of neural networks and attract the D2V encoder during training under the influence of $\mathcal{L}_{en}$. 

\subsection{Class-Adversarial Training on D2V}
\label{sec:class_adv}
We expect the D2V encoder $h_v$ to contain as little class-specific information as possible, which ensures that HMOE partitions the input space based on domain-wise distinction rather than semantic categories. Inspired by Domain-Adversarial Neural Networks \cite{ganinDomainadversarialTrainingNeural2016}, we define an adversarial classifier $f_c^{ad}$ taking $v$ as input and add the following loss to perform class-adversarial training on $h_v$:
\begin{equation}
    \mathcal{L}_{ad} = \mathbb{E}_{(x, y) \sim \mathcal{D}_{tr}} \left[ \ell_{ce}(f_c^{ad}(GRL(v, \lambda_{grl})), y) \right]
\end{equation}
where $\ell_{ce}$ denotes the cross-entropy loss and $GRL$ represents the gradient reversal layer, which acts as an identity function in the forward pass and multiplies the gradient by $-\lambda_{grl}$ in the backward pass. As suggested in \cite{ganinDomainadversarialTrainingNeural2016}, we define $\lambda_{grl}$ as follows:
\begin{equation}
	 \lambda_{grl} = 2 / (1 + \exp (-10 \times pct_{tr})) - 1
\end{equation}
where $pct_{tr}$ varies linearly from 0 to 1 during training.

\subsection{Supervised Learning on Targets}

We provide two ways to calculate the supervised loss on targets $\mathcal{L}_y$, that is, Empirical Risk Minimization (ERM) \cite{vapnikNatureStatisticalLearning1999} and the intra-domain \emph{mixup}.

\vspace{0.2em}
\noindent \textbf{ERM} \quad In the setting of ERM, the supervised loss on targets is simply the empirical risk on the training data $\mathcal{D}_{tr}$:
\begin{equation}
	\mathcal{L}_{y} = \mathbb{E}_{(x, y) \sim \mathcal{D}_{tr}} \left[ \ell_{ce}(\hat{y}, y) \right]  \label{eq:erm}
\end{equation}
where $\hat{y}$ is the prediction of HMOE, as calculated by \cref{eq:output}.

\vspace{0.2em}
\noindent \textbf{Intra-domain mixup} \quad \emph{mixup} trains a neural network on virtual samples synthesized through convex combinations of pairs of samples and their labels \cite{zhangMixupEmpiricalRisk2017}:
\begin{align}
	\tilde{x} &= \beta x_i + (1 - \beta) x_j  \\
	\tilde{y} &= \beta y_i + (1 - \beta) y_j
\end{align}
where $\beta \sim \text{Beta}(\alpha, \alpha)$ and $\alpha$ adjusts interpolation strength. \emph{mixup} can be seen as a data augmentation approach theoretically grounded in Vicinal Risk Minimization \cite{chapelleVicinalRiskMinimization2000}, which is an alternative learning principle to ERM. \cite{xuAdversarialDomainAdaptation2020, yanImproveUnsupervisedDomain2020} applied the inter-domain \emph{mixup} mixing samples across different domains for domain-invariant learning, whereas our intra-domain \emph{mixup}, as shown in \cref{alg:mixup}, prompts HMOE for smoother predictions in neighborhood within each domain, enhancing its generalization and robustness.

To perform the intra-domain \emph{mixup} without domain labels, HMOE starts with \cref{eq:erm} and then switches to \cref{alg:mixup} until $\mathcal{L}_{en} < 0.1$ indicating latent domains are reasonably discovered and clustered.

\begin{algorithm}[htbp]
	\caption{intra-domain \emph{mixup}}
    \label{alg:mixup}
	\begin{algorithmic}[1]
		\Require A mini-batch $\mathcal{B}$ split into distinct domains given domain labels or clusters identified by gate values
		\ForEach {domain or cluster $\mathcal{B}_i \in \mathcal{B}$}
			\State $\widetilde{\mathcal{B}_i} \gets \emph{mixup}(\mathcal{B}_i, \text{shuffled } \mathcal{B}_i)$ with $\beta \sim \text{Beta}(\alpha, \alpha)$
			\Comment{\textcolor{cyan}{Mix same-index samples between $\mathcal{B}_i$ and $\text{shuffled } \mathcal{B}_i$}}
			\State Compute the empirical risk $\mathcal{L}_i$ on $\widetilde{\mathcal{B}_i}$
		\EndFor
		\State $\mathcal{L}_{y} \gets $ Average over all $\mathcal{L}_i$
	\end{algorithmic}
\end{algorithm}

\subsection{Semi-/supervised Learning on Domains}
\label{sec:semi_domains}

Due to the probabilistic nature of MoE, given an input $x$ and the corresponding gate values $\bm{p} = \{ p_k \}_{k=1}^K$, we can interpret $p_k$ as the probability of selecting the $k$th expert $E_k$ given $x$, \ie, $p(E_k | x)$.  In addition, $E_k$ is thought to be associated with a specific domain $\mathcal{S}_m$. Therefore, we get $p_k = p(E_k | x) = p(S_m | x)$. Consider a dataset with domain labels $\mathcal{D}_d = \{ (x_i, d_i) \}_{i=1}^{N_d}$ (class labels are not necessary) with $d_i \in \{ 1, \ldots, M_d \}$, we can make use of $\mathcal{D}_d$ as follows:
\begin{equation}
    \mathcal{L}_{d} = \mathbb{E}_{(x, d) \sim \mathcal{D}_d} \left[ \ell_{ce}(\bm{p}, d) \right]
\end{equation}
$M_d$ may be smaller than $K$, but this has no bearing on the calculation of $\mathcal{L}_{d}$. In this case, we assume that the first $M_d$ experts are assigned to $M_d$ domains, while the rest learn autonomously without domain information.  If all domain labels are given, $\mathcal{L}_{d}$ shifts to supervised domain learning.

\subsection{Training and Inference}
\label{sec:train_infer}

The final training loss is:
\begin{equation}
    \mathcal{L} = \lambda_{y} \mathcal{L}_{y} + \lambda_{en} \mathcal{L}_{en} + \lambda_{kl} \mathcal{L}_{kl} + \lambda_{ad} \mathcal{L}_{ad} + \lambda_{d} \mathcal{L}_{d}
\end{equation}
where $\lambda$ are trade-off hyper-parameters to balance different losses. Generally, $\lambda_{y}$ is set to 1 and $\mathcal{L}_{d}$ is not used for compound DG without domain labels.

For inference, we offer two modes: MIX and OOD. MIX refers to the mixture of experts, as calculated by \cref{eq:output}. OOD\footnote{OOD is efficiently realized using PyTorch-based JAX-like \emph{functorch}.} (Out of Domain) uses the output of a classifier whose weights are generated by the hypernetwork directly taking the D2V encoder as input. OOD enables the generalization of experts beyond aggregation.

\section{Experiments}
\label{sec:exp}

This paper focuses on image classification. However, to illustrate HMOE's learning dynamics and versatility, we also apply it to a toy regression task to learn a one-dimensional function defined on 3 intervals. HMOE proves effective in assigning an expert to each interval. Due to space limits, details are in the \textbf{supplementary material}. Next, we evaluate HMOE against other DG algorithms on DomainBed \cite{gulrajaniSearchLostDomain2020}.

\subsection{Datasets and Model Evaluation}
DomainBed offers a unified codebase to implement, train, and evaluate DG algorithms, and integrates commonly used DG-related datasets. We experiment on Colored MNIST (3 domains and 2 classes) \cite{arjovskyInvariantRiskMinimization2019}, Rotated MNIST (6 domains and 10 classes) \cite{ghifaryDomainGeneralizationObject2015},  PACS (4 domains and 7 classes) \cite{liDeeperBroaderArtier2017}, VLCS (4 domains and 5 classes) \cite{fangUnbiasedMetricLearning2013}, OfficeHome (4 domains and 65 classes) \cite{venkateswaraDeepHashingNetwork2017}, and TerraIncognita (4 domains and 10 classes) \cite{beeryRecognitionTerraIncognita2018}. Detailed dataset statistics and sample visualization are provided in the \textbf{supplementary material}.

For model selection and hyper-parameter tuning, DomainBed offers three options, of which we choose the training-domain validation that allocates 80\% from each training domain for training and the rest for validation. This option aligns well with compound DG without access to domain labels and test domains.

\subsection{Implementation Details}

For CMNIST and RMNIST, we use a four-layer ConvNet as the featurizer (see Appendix D.1 of \cite{gulrajaniSearchLostDomain2020}). The D2V encoder $h_v$ connects this four-layer ConvNet to a fully-connected (fc) layer in order to map to the embedding dimension $D$.

For other datasets, we use ResNet-50 pretrained on ImageNet \cite{dengImagenetLargescaleHierarchical2009} as the featurizer and freeze all batch normalization layers. The D2V encoder $h_v$ cascades 3 \emph{conv} layers (64-128-256 units, stride 2, $4 \times 4$ kernels, ReLU), two residual blocks (each has 2 \emph{conv} layers with 256 units, $3 \times 3$ kernels, ReLU), and a  $3 \times 3$ \emph{conv} layer with $D$ units followed by global average pooling. We use Instance Normalization \cite{ulyanovInstanceNormalizationMissing2016} with learnable affine parameters before all ReLU of $h_v$.

For all datasets, the classifier $f_c$ is a fc layer whose input size is the featurizer's output size (128 for ConvNet and 2048 for ResNet-50) and output size is the number of classes. The hypernetwork $f_h$ is a five-layer MLP with 256-128-64-32 hidden units and SiLU \cite{hendrycksGaussianErrorLinear2016}, and its input size is $D$ and output size is the total number of learnable parameters (\ie, weights and biases) of $f_c$.  In addition, we initialize $f_h$ using the hyperfan method \cite{changPrincipledWeightInitialization2019}. If $\mathcal{L}_{ad}$ is used, the adversarial classifier is a three-layer MLP with 256 hidden units and ReLU, and its input size is $D$ and output size is the number of classes. We set $D=32$ and initialize embedding vectors with the standard normal distribution.

We define three HMOE variants, including
(1) \textbf{HMOE-DL}:  Domain labels are provided. We use $\mathcal{L}_{y}$ calculated by \cref{eq:erm} and $\mathcal{L}_{d}$ with $\lambda_{y} = \lambda_{d} = 1$ and discard other losses, and $K$ is the number of training domains.
(2) \textbf{HMOE-ND}: No domain information is available. We use $\mathcal{L}_{y}$ calculated by \cref{eq:erm}, $\mathcal{L}_{en}$, $\mathcal{L}_{kl}$ and $\mathcal{L}_{ad}$ with $\lambda_{y} = \lambda_{en} = \lambda_{kl} = 1$ and $\lambda_{ad} = 0.1$, and we fix $K=3$.
(3) \textbf{HMOE-MU}: The setting is the same as in HMOE-ND, except that $\mathcal{L}_{y}$ is calculated via the intra-domain \emph{mixup} (\cref{alg:mixup}) with $\alpha = 0.3$.

DomainBed trains all DG algorithms with Adam for 5,000 iterations. For Colored and Rotated MNIST / other datasets, the learning rate is 0.001 / 5e-5, the batch size is 64 / 32 $\times$ the number of training domains, and models are evaluated on the validation set every 100 / 300 iterations. Each experiment uses one domain of a dataset as the test domain and trains algorithms on the others, which is repeated three times with different random seeds. The average accuracy over three replicates is reported. DG algorithms use the default settings predefined in DomainBed. All experiments are conducted using PyTorch on multiple A5000 GPUs.

\subsection{Results}

The DomainBed benchmark in \cite{gulrajaniSearchLostDomain2020} has been outdated, and we update it using an improved pretrained ResNet-50 (IMAGENET1K-V2) available on torchvision. The comparison of HMOE against other DG algorithms is shown in \cref{tab:domainbed}, where DeepAll means the vanilla supervised learning that just fine-tunes ResNet-50 on mixed data and serves as a performance baseline. We report the average accuracy of all test domains for each dataset. Refer to the \textbf{supplementary material} for detailed results.

\begin{table*}[htbp]
        \centering
		\small
		\setlength\tabcolsep{5pt}
        \begin{tabular}{lccccccccc|c}
        \hline
        \textbf{Algorithm}  & \emph{\textbf{M}} & & \textbf{CMNIST}     & \textbf{RMNIST}     & \textbf{VLCS}             & \textbf{PACS}             & \textbf{OfficeHome}       & \textbf{TerraInc}  & \textbf{Avg}& \textbf{Ranking} \\
        \hline
        \multicolumn{11}{c}{\emph{w/ Domain Labels}} \\
        \hline
		
		Mixup         \cite{yanImproveUnsupervisedDomain2020}                 & \checkmark & & 51.9 {\scriptsize$\pm$ 0.1} & \textcolor{gray}{97.6} {\scriptsize$\pm$ 0.1}& 78.7 {\scriptsize$\pm$ 0.1} 								 & 86.6 {\scriptsize$\pm$ 0.1} 								 & 71.6 {\scriptsize$\pm$ 0.2} 								  & 51.4 {\scriptsize$\pm$ 0.4} & \underline{72.97} & \underline{2}\\
		CORAL       \cite{sunDeepCoralCorrelation2016}                               & & 						& 51.4 {\scriptsize$\pm$ 0.1} & 98.0 {\scriptsize$\pm$ 0.0} & 78.1 {\scriptsize$\pm$ 0.2} 							     & 86.7 {\scriptsize$\pm$ 0.4} 								& \underline{72.2} {\scriptsize$\pm$ 0.2} 								 & 48.9 {\scriptsize$\pm$ 0.5} & 72.55 & 5\\
		VREx         \cite{kruegerOutofdistributionGeneralizationRisk2021}  & & 					  & \underline{52.2} {\scriptsize$\pm$ 0.1} & 97.8 {\scriptsize$\pm$ 0.0} & \textcolor{gray}{77.3} {\scriptsize$\pm$ 0.2}  & 86.0 {\scriptsize$\pm$ 0.7} 						       & 69.8 {\scriptsize$\pm$ 0.1} 								& 51.8 {\scriptsize$\pm$ 0.4} & 72.48 & 6\\
		Fish           \cite{shiGradientMatchingDomain2021a}                         & & 					   & 51.5 {\scriptsize$\pm$ 0.1} & 97.9 {\scriptsize$\pm$ 0.1} & 78.1 {\scriptsize$\pm$ 0.0} 								 & 86.9 {\scriptsize$\pm$ 0.9} 								 & 68.7 {\scriptsize$\pm$ 0.1} 								  & 51.0 {\scriptsize$\pm$ 0.7} & 72.35   & 7\\
		ARM          \cite{zhangAdaptiveRiskMinimization2021}                      & & 						& \textcolor{cyan}{55.6} {\scriptsize$\pm$ 0.3} & \textcolor{cyan}{98.1} {\scriptsize$\pm$ 0.0} & 78.0 {\scriptsize$\pm$ 0.6} 								& \textcolor{gray}{85.7} {\scriptsize$\pm$ 0.8} &  \textcolor{gray}{66.5} {\scriptsize$\pm$ 0.4} & 48.5 {\scriptsize$\pm$ 0.4} & 72.05 & 9\\
		MTL            \cite{blanchardDomainGeneralizationMarginal2021}     & & 					  & 51.5 {\scriptsize$\pm$ 0.2} & 97.8 {\scriptsize$\pm$ 0.0} & \textcolor{gray}{77.3} {\scriptsize$\pm$ 0.3}   & \textcolor{gray}{85.5} {\scriptsize$\pm$ 0.2} &  \textcolor{gray}{68.4} {\scriptsize$\pm$ 0.5} & 51.3 {\scriptsize$\pm$ 0.6} & 71.97 & 10\\
		GroupDRO \cite{sagawaDistributionallyRobustNeural2020}            & & 						& 52.1 {\scriptsize$\pm$ 0.0} & 97.8 {\scriptsize$\pm$ 0.0} & 77.8 {\scriptsize$\pm$ 0.6} 								 & \textcolor{gray}{85.0} {\scriptsize$\pm$ 0.8} &  \textcolor{gray}{68.3} {\scriptsize$\pm$ 0.3} & 49.6 {\scriptsize$\pm$ 0.5} & 71.77 & 11\\
		MLDG         \cite{liLearningGeneralizeMetalearning2018}                 & & 						& \textcolor{gray}{44.2} {\scriptsize$\pm$ 4.6}& 97.8 {\scriptsize$\pm$ 0.0} & \textcolor{gray}{76.6} {\scriptsize$\pm$ 0.2} & \underline{87.1} {\scriptsize$\pm$ 0.1} 							    &  \textcolor{gray}{68.3} {\scriptsize$\pm$ 0.3} & 49.9 {\scriptsize$\pm$ 1.1} & \textcolor{gray}{70.65} & \textcolor{gray}{15}\\
		MMD          \cite{liDomainGeneralizationAdversarial2018}                & & 						& \textcolor{gray}{38.5} {\scriptsize$\pm$ 0.8}& 98.0 {\scriptsize$\pm$ 0.0} & \textcolor{gray}{77.4} {\scriptsize$\pm$ 0.9} & \textcolor{gray}{84.2} {\scriptsize$\pm$ 0.1} & 69.1 {\scriptsize$\pm$ 0.0} 								 & 50.0 {\scriptsize$\pm$ 1.2} & \textcolor{gray}{69.53} & \textcolor{gray}{16}\\
		DANN        \cite{ganinDomainadversarialTrainingNeural2016}        & & 					 & 51.8 {\scriptsize$\pm$ 0.1} & \textcolor{gray}{97.7} {\scriptsize$\pm$ 0.0}& \textcolor{gray}{75.6} {\scriptsize$\pm$ 0.6}   & \textcolor{gray}{77.0} {\scriptsize$\pm$ 1.4}  &  \textcolor{gray}{66.5} {\scriptsize$\pm$ 0.3} &  \textcolor{gray}{42.5} {\scriptsize$\pm$ 2.6} & \textcolor{gray}{68.52} & \textcolor{gray}{17}\\
        IRM             \cite{arjovskyInvariantRiskMinimization2019}                 & &  					   & \textcolor{gray}{41.3} {\scriptsize$\pm$ 0.9}& \textcolor{gray}{87.3} {\scriptsize$\pm$ 0.4}& 78.3 {\scriptsize$\pm$ 1.1} 								& \textcolor{gray}{82.1} {\scriptsize$\pm$ 0.7}  &  \textcolor{gray}{64.9} {\scriptsize$\pm$ 0.3} & 50.8 {\scriptsize$\pm$ 1.1} & \textcolor{gray}{67.45} & \textcolor{gray}{18}\\     
        \cdashline{1-10}
        \multirow{2}{5em}{HMOE-DL}                                   				  &      & MIX & 51.5 {\scriptsize$\pm$ 0.1} & \textcolor{gray}{94.1} {\scriptsize$\pm$ 0.5}	& \textcolor{gray}{77.0} {\scriptsize$\pm$ 0.4}  & \textcolor{gray}{85.5} {\scriptsize$\pm$ 0.6}  &  68.9 {\scriptsize$\pm$ 0.6} 				              & 49.6 {\scriptsize$\pm$ 0.2} &  \textcolor{gray}{71.70}  &   \textcolor{gray}{14}\\
                                                                     			   &      & \cellcolor{middlegrey} OOD & \cellcolor{middlegrey} 57.0 {\scriptsize$\pm$ 3.9} & \cellcolor{middlegrey} 93.3 {\scriptsize$\pm$ 0.5} & \cellcolor{middlegrey} 77.9 {\scriptsize$\pm$ 0.3} & \cellcolor{middlegrey} 85.1 {\scriptsize$\pm$ 0.8} & \cellcolor{middlegrey} 67.9 {\scriptsize$\pm$ 0.3} & \cellcolor{middlegrey} 48.3 {\scriptsize$\pm$ 0.4} & \cellcolor{middlegrey} 71.58   &  \\
        \hline
        \multicolumn{11}{c}{\emph{w/o Domain Labels}}  \\
        \hline
        SelfReg     \cite{kimSelfregSelfsupervisedContrastive2021}            & \checkmark & & 51.4 {\scriptsize$\pm$ 0.1} & 98.0 {\scriptsize$\pm$ 0.0} & \textcolor{cyan}{78.9} {\scriptsize$\pm$ 0.3} 								  & 86.1 {\scriptsize$\pm$ 0.3} 							  & 71.3 {\scriptsize$\pm$ 0.2}    								& 51.5 {\scriptsize$\pm$ 0.3} & 72.87    & 3\\
        SagNet    \cite{namReducingDomainGap2021}                                 &  & 				       & 51.8 {\scriptsize$\pm$ 0.1} & 98.0 {\scriptsize$\pm$ 0.0} & 77.7 {\scriptsize$\pm$ 0.3} 								 & 86.2 {\scriptsize$\pm$ 0.4} 						        & 69.3 {\scriptsize$\pm$ 0.2}    							 & 50.7 {\scriptsize$\pm$ 0.5} & 72.28    & 8\\
        RSC          \cite{huangSelfchallengingImprovesCrossdomain2020} &  & 				    & 51.5 {\scriptsize$\pm$ 0.2} & \textcolor{gray}{97.5} {\scriptsize$\pm$ 0.1}& \underline{78.8} {\scriptsize$\pm$ 0.3} 							      & 87.0 {\scriptsize$\pm$ 0.4} 						     &  \textcolor{gray}{65.5} {\scriptsize$\pm$ 0.9}  & 49.1 {\scriptsize$\pm$ 1.0} & 71.57     & 12\\
        DeepAll         \cite{vapnikNatureStatisticalLearning1999}                      &  & 					 & 51.4 {\scriptsize$\pm$ 0.1} & 97.8 {\scriptsize$\pm$ 0.1} & 77.5 {\scriptsize$\pm$ 0.2} 								   & 85.8 {\scriptsize$\pm$ 0.4} 							  & 68.5 {\scriptsize$\pm$ 0.2}    								& 47.7 {\scriptsize$\pm$ 0.9} & 71.45   & 13\\
        \cdashline{1-10}
        \multirow{2}{5em}{HMOE-ND}                                   			    &       & MIX & 				  51.8 {\scriptsize$\pm$ 0.1} & \textcolor{gray}{97.5} {\scriptsize$\pm$ 0.1} & 78.1 {\scriptsize$\pm$ 0.3} 									& 86.6 {\scriptsize$\pm$ 0.3} 							   & 69.7 {\scriptsize$\pm$ 0.2}  								& \underline{52.5} {\scriptsize$\pm$ 0.3} &   72.70   & 4\\
        										                     	    &    & \cellcolor{middlegrey} OOD & \cellcolor{middlegrey} 51.8 {\scriptsize$\pm$ 0.1} & \cellcolor{middlegrey} 97.5 {\scriptsize$\pm$ 0.1} & \cellcolor{middlegrey} 78.0 {\scriptsize$\pm$ 0.4} & \cellcolor{middlegrey} 86.9 {\scriptsize$\pm$ 0.2} & \cellcolor{middlegrey} 69.0 {\scriptsize$\pm$ 0.2} & \cellcolor{middlegrey} 51.1 {\scriptsize$\pm$ 1.4} & \cellcolor{middlegrey}  72.38     &  \\
        \cdashline{1-10}
        \multirow{2}{5em}{HMOE-MU}                                                  & \multirow{2}{1em}{\checkmark} & MIX  & 51.7 {\scriptsize$\pm$ 0.2} & \textcolor{gray}{97.6} {\scriptsize$\pm$ 0.1} & 78.6 {\scriptsize$\pm$ 0.0} 								  & \textcolor{cyan}{88.0} {\scriptsize$\pm$ 0.3} 							   & \textcolor{cyan}{72.5} {\scriptsize$\pm$ 0.1}  							  & \textcolor{cyan}{52.8} {\scriptsize$\pm$ 0.9} &   \textcolor{cyan}{73.53}    &  \textcolor{cyan}{1}\\
           	                                                        						                  & &  \cellcolor{middlegrey} OOD & \cellcolor{middlegrey} 51.6 {\scriptsize$\pm$ 0.2} & \cellcolor{middlegrey} 97.6 {\scriptsize$\pm$ 0.1} & \cellcolor{middlegrey} 78.8 {\scriptsize$\pm$ 0.3} & \cellcolor{middlegrey} 87.0 {\scriptsize$\pm$ 1.0} & \cellcolor{middlegrey} 72.4 {\scriptsize$\pm$ 0.1} & \cellcolor{middlegrey} 52.1 {\scriptsize$\pm$ 0.9} & \cellcolor{middlegrey} 73.25     & \\
        \hline
        \end{tabular}
        \caption{Domain generalization results on DomainBed. We format \textcolor{cyan}{first}, \underline{second} and \textcolor{gray}{worse than DeepAll} results. \emph{M} denotes the inter/intra-\emph{mixup}. The performance of HMOE is evidenced by MIX. OOD is only for comparison with MIX and does not participate in the ranking. }
        \label{tab:domainbed}
\end{table*}

HMOE-MU outperforms all other DG algorithms in average accuracy. Notably, \emph{mixup}-powered algorithms show impressive performance, proving the effectiveness of \emph{mixup} in enhancing generalization. Both Mixup \cite{yanImproveUnsupervisedDomain2020} (second place) and SelfReg \cite{kimSelfregSelfsupervisedContrastive2021} (third place) adopted the inter-domain \emph{mixup} to learn domain-invariant representations. 
HMOE-ND ranks fourth overall, but is the top among algorithms without \emph{mixup}. In addition, HMOE-ND / MU largely surpass the DeepAll baseline, except on RMNIST.

For MNIST datasets, performance is comparable across algorithms, except for the outstanding results of ARM \cite{zhangAdaptiveRiskMinimization2021}. Other datasets pose higher challenges. For instance, VLCS comprises real photo images, with the domain shift primarily caused by changes in scene and perspective, leading to subtle visual differences between domains. Many algorithms are inferior to DeepAll on these challenging datasets. HMOE-MU achieves state-of-the-art results on PACS, OfficeHome, and TerraInc, and its performance on VLCS is nearly on par with the best result (78.6 \emph{vs}. 78.9). HMOE-ND also performs impressively. All these findings validate the superiority of HMOE in addressing compound DG.

\begin{figure}[htbp]
    \centering
    \begin{subfigure}{0.55\linewidth}
		\includegraphics[width=\textwidth]{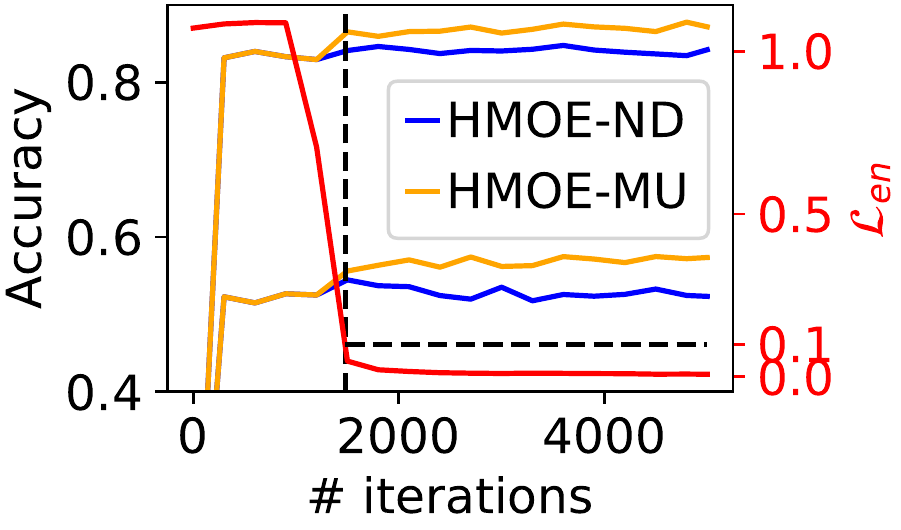}
		\caption{HMOE-ND \emph{vs.} MU}
		\label{fig:nd_vs_mixup}
	\end{subfigure}
    \hspace{0.2em}
    \begin{subfigure}{0.4\linewidth}
        \includegraphics[width=\textwidth]{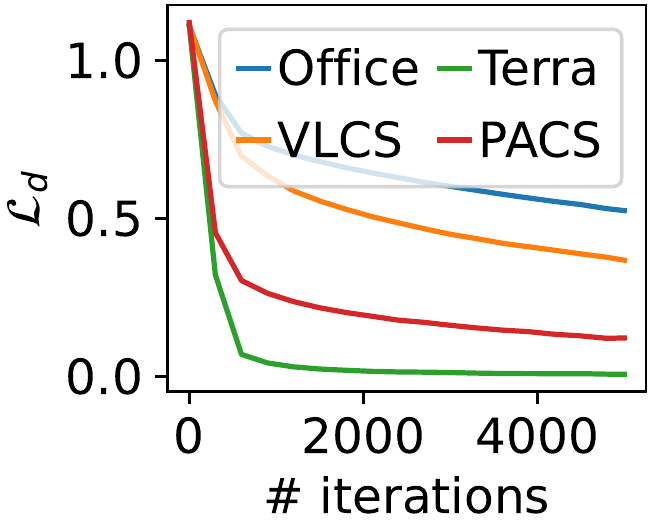}
        \caption{Average of $\mathcal{L}_d$ per dataset}
        \label{fig:hmoe_dl_ld}
    \end{subfigure}
    \caption{Losses over iterations. (a) ND / MU on OfficeHome with \emph{clipart} as the test domain. The upper / lower curves of ND / MU represent their validation / test accuracy, respectively.}
\end{figure}

HMOE-MU markedly surpasses ND. \cref{fig:nd_vs_mixup} presents a comparison of their validation / test accuracy during training. It is evident that the accuracy of MU continues to improve with the introduction of intra-domain \emph{mixup} upon $\mathcal{L}_{en} < 0.1$, because \emph{mixup} imposes linearity constraints, which prompts smoothness and mitigates overfitting.

Interestingly, HMOE-DL lags behind HMOE-ND / MU significantly, indicating that HMOE performs better when using self-learned domain information rather than relying on provided domain labels. We observe that the latent domains discovered by HMOE seem to be more human-intuitive than given domain labels (\cref{sec:domain_discovery}). \cref{fig:hmoe_dl_ld} shows that the supervised loss on domains $\mathcal{L}_d$ of HMOE-DL fails to decrease rapidly on OfficeHome and VLCS datasets. This could suggest that HMOE struggles to assimilate domain label information, which complicates its learning process and negatively affects its DG performance.

For two inference modes, MIX outperforms OOD in most cases, but OOD can be used to sacrifice a little accuracy for efficiency in practice because it is more computationally efficient without computing all experts like MIX.

\subsection{Latent Domain Discovery}
\label{sec:domain_discovery}

\begin{figure*}
    \centering
    \begin{subfigure}{0.158\linewidth}
        \includegraphics[width=\textwidth]{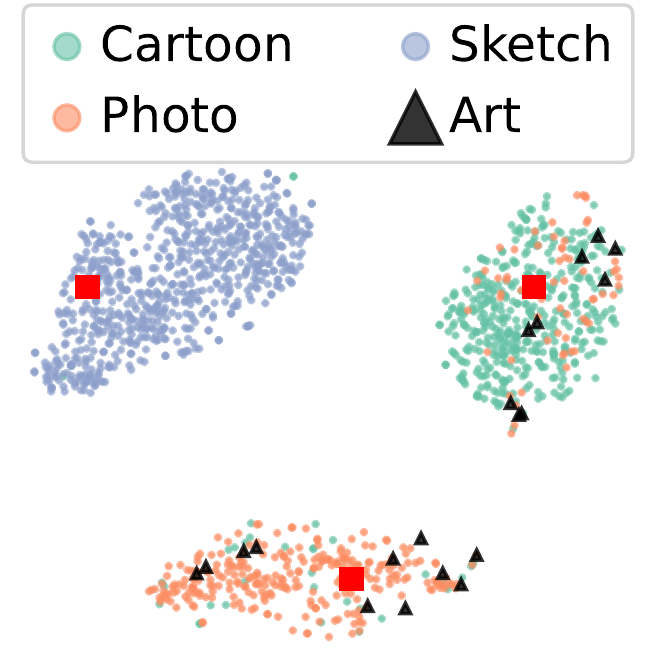}
        \caption{PACS-ND (Art)}
        \label{fig:pacs_art_ND}
    \end{subfigure}
    \begin{subfigure}{0.1625\linewidth}
        \includegraphics[width=\textwidth]{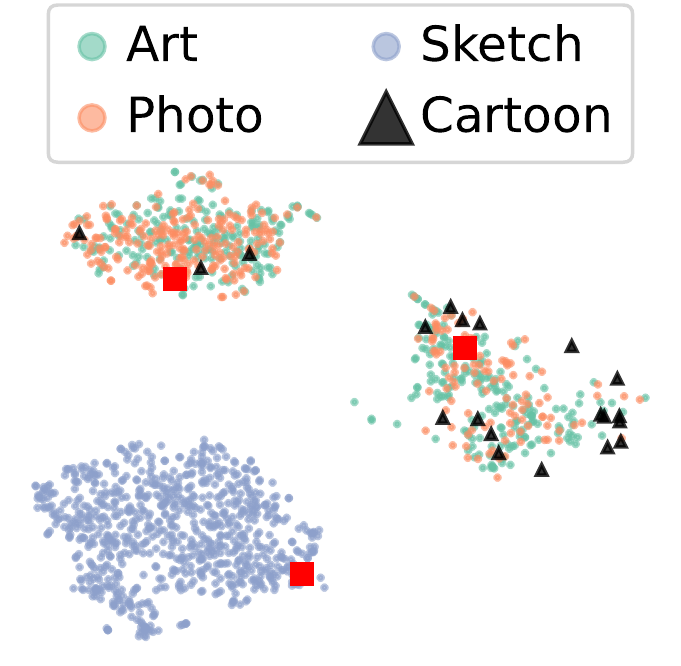}
        \caption{PACS-ND (Cartoon)}
        \label{fig:pacs_cartoon_ND}
    \end{subfigure}
    \begin{subfigure}{0.1625\linewidth}
        \includegraphics[width=\textwidth]{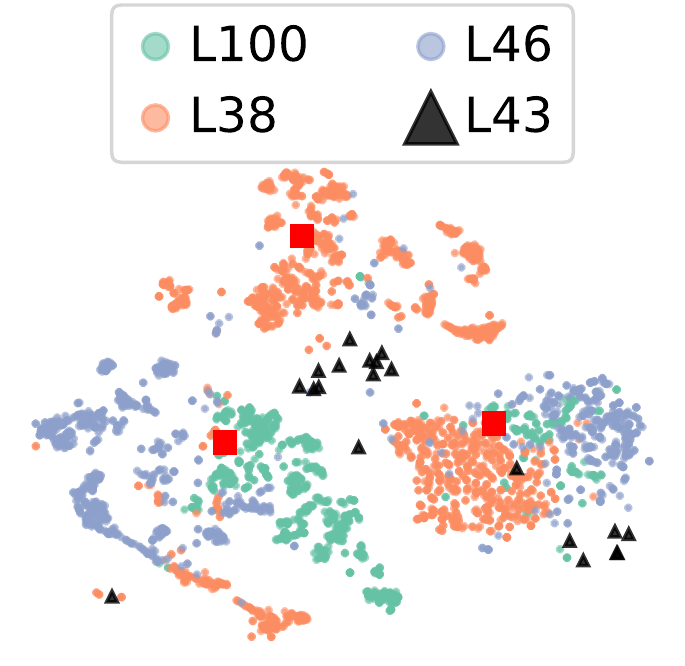}
        \caption{TerraInc-ND (L43)}
        \label{fig:terra_L32_ND}
    \end{subfigure}
    \begin{subfigure}{0.16\linewidth}
        \includegraphics[width=\textwidth]{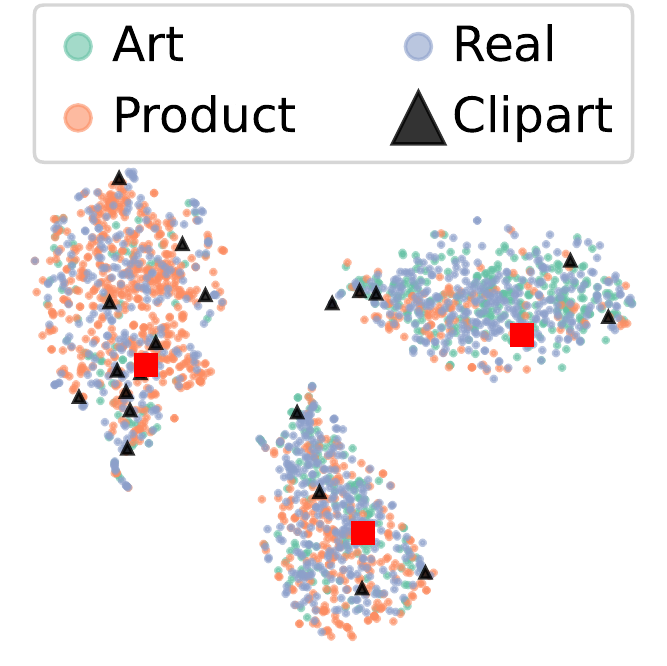}
        \caption{Office-ND (Clipart)}
        \label{fig:office_clipart_ND}
    \end{subfigure}
    \begin{subfigure}{0.1625\linewidth}
        \includegraphics[width=\textwidth]{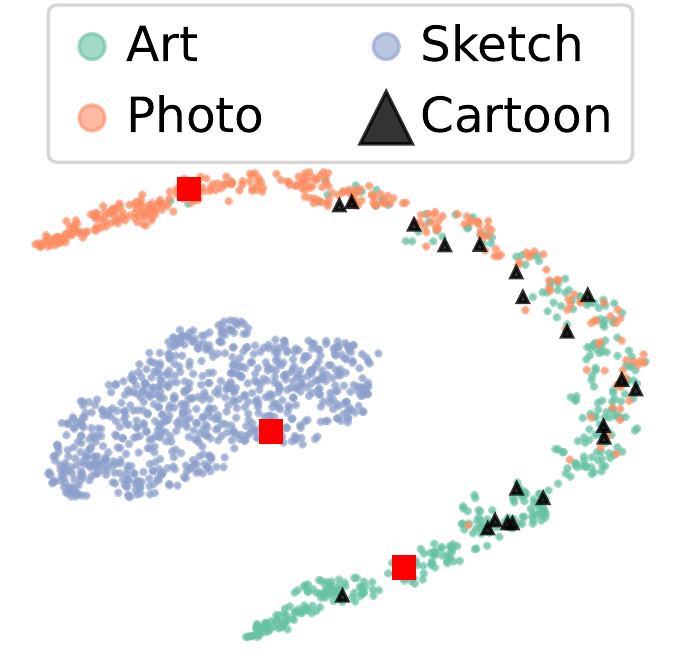}
        \caption{PACS-DL (Cartoon)}
        \label{fig:pacs_cartoon_DL}
    \end{subfigure}
    \begin{subfigure}{0.16\linewidth}
        \includegraphics[width=\textwidth]{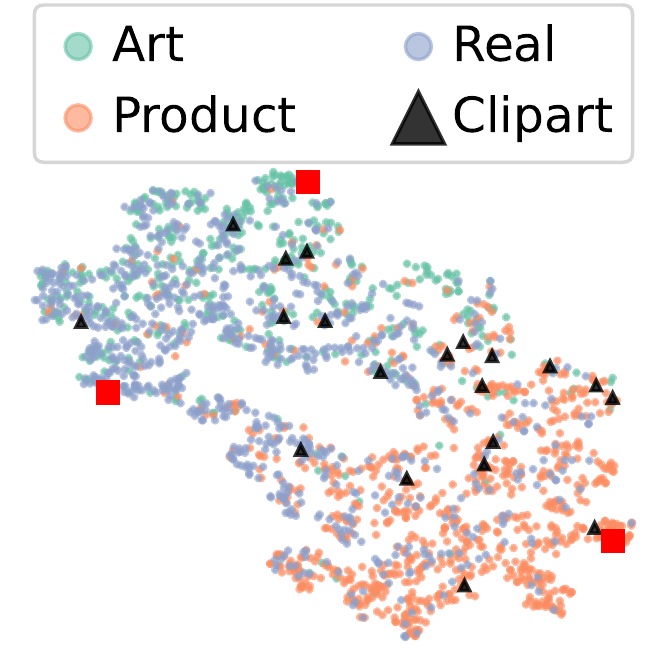}
        \caption{Office-DL (Clipart)}
        \label{fig:office_clipart_DL}
    \end{subfigure}
 
    \caption{The t-SNE visualization of the output of the D2V encoder. The suffixes in captions denote HMOE-DL / ND, with the test domain in parentheses. Red squares are embedding vectors, black triangles are 20 samples randomly drawn from the test domain, and other dots are training domains. The silhouette coefficients are 0.7, 0.68, 0.48, 0.66, 0.46, and 0.36 for \cref{fig:pacs_cartoon_DL,fig:office_clipart_DL,fig:pacs_art_ND,fig:pacs_cartoon_ND,fig:terra_L32_ND,fig:office_clipart_ND}, respectively.}
    \label{fig:t-SNE}
\end{figure*}

We employ t-SNE \cite{vandermaatenVisualizingDataUsing2008} to visualize the output of the D2V encoder, as shown in \cref{fig:t-SNE}. It is evident that HMOE-ND effectively separates the mixed data into distinct clusters, each gravitating towards an embedding vector. 

Domain labels are used to color data to highlight the differences between them and inferred latent domains. For PACS with the \emph{art} test (\cref{fig:pacs_art_ND}), inferred domains largely align with domain labels, although some photos are grouped into the cartoon-predominant cluster. However, with \emph{cartoon} as the test domain (\cref{fig:pacs_cartoon_ND}), data is not split based on \emph{art} and \emph{photo}. \cref{fig:pacs_cartoon_DL} shows that, even with domain labels, HMOE-DL struggles to fully separate \emph{art} from \emph{photo}. For TerraInc (\cref{fig:terra_L32_ND}), points of the same color tend to cluster together, whereas for OfficeHome (\cref{fig:office_clipart_ND}), different colors intermix within each cluster, highlighting the big gap between labeled and inferred domains. \cref{fig:office_clipart_DL} also shows that HMOE-DL has difficulty in data partitioning, explaining the slow decrease in $\mathcal{L}_d$ for OfficeHome in \cref{fig:hmoe_dl_ld}.

\begin{figure}[htbp]
	\centering
    \begin{subfigure}{0.495\linewidth}
		\includegraphics[width=\textwidth]{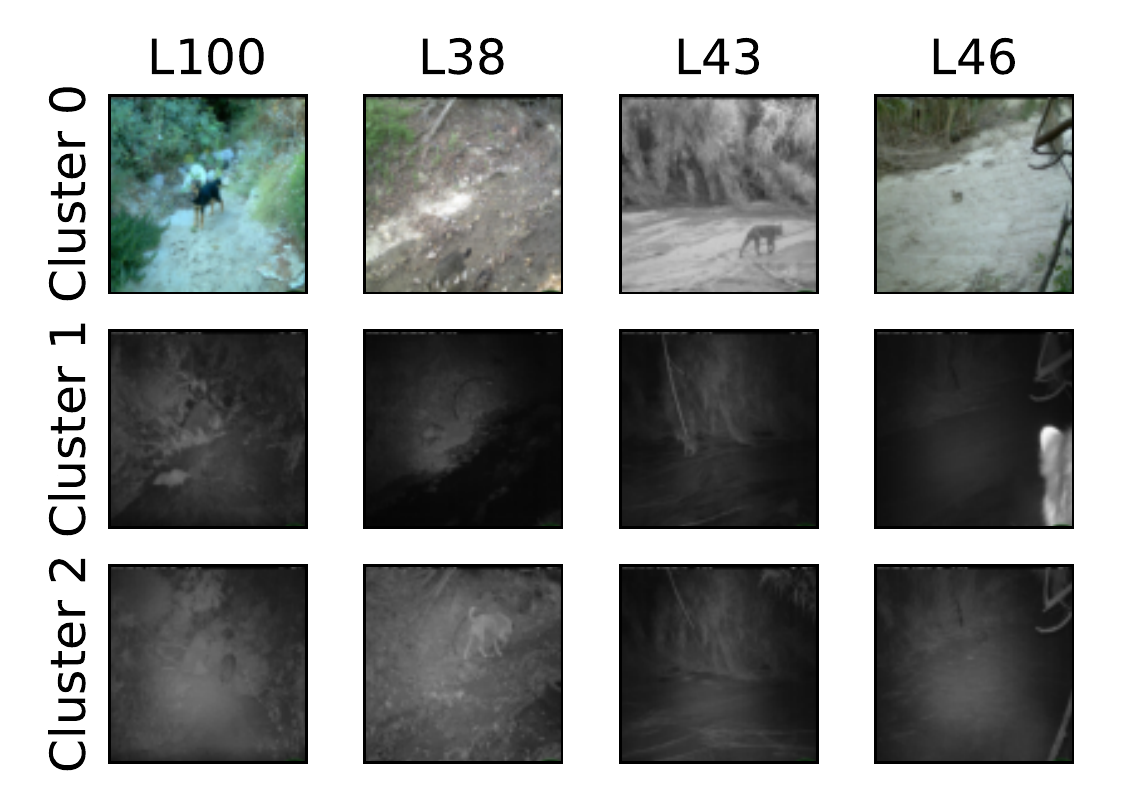}
		\caption{TerraInc-ND (L43)}
		\label{fig:terra_domains}
	\end{subfigure}
	\begin{subfigure}{0.495\linewidth}
		\includegraphics[width=\textwidth]{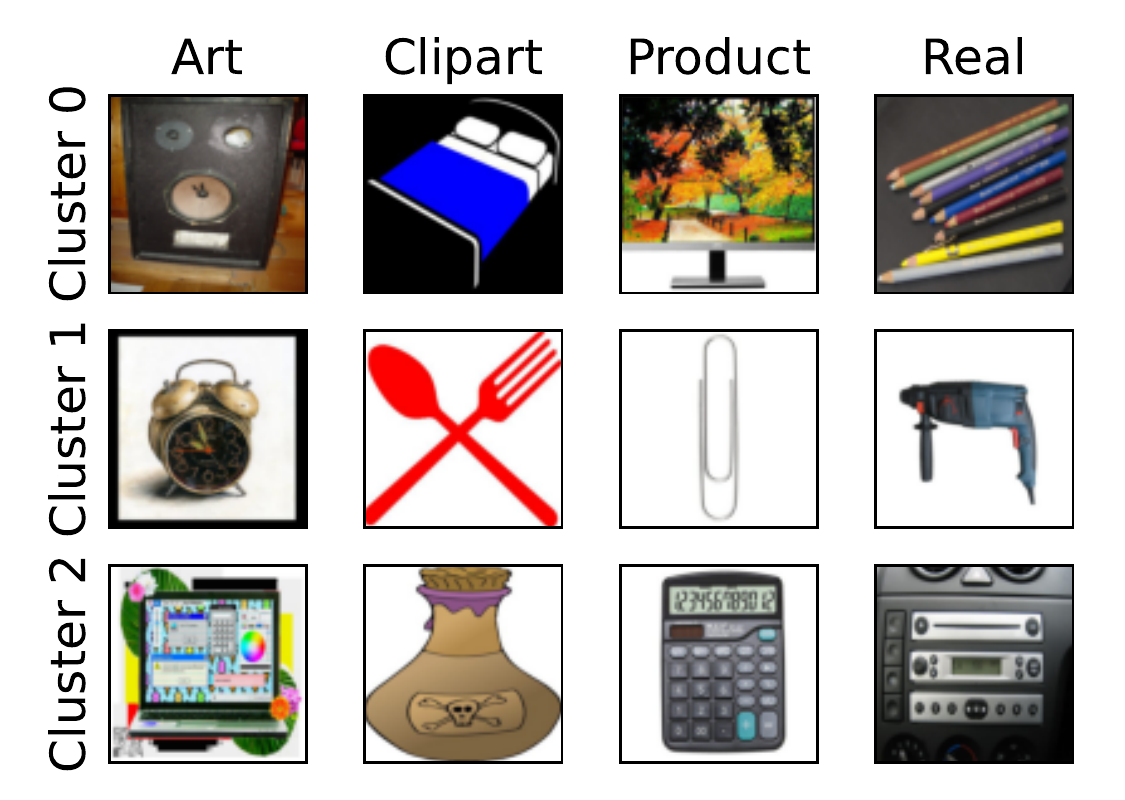}
		\caption{Office-ND (Clipart)}
		\label{fig:office_domains}
	\end{subfigure}
    
	\caption{Compare domain labels and HMOE-ND clusters}
	\label{fig:cluster_imgs}
\end{figure}

To intuitively understand how HMOE distinguishes between domains, \cref{fig:cluster_imgs} compares labeled and inferred domains using visual samples. HMOE-ND seems to partition TerraInc by illumination and OfficeHome by background complexity, which aligns more with human intuition.

After the above analysis, we conclude that the success of HMOE stems from its ability to self-learn more reasonable domain knowledge. However, this does not mean that given domain labels are erroneous. There are typically multiple generative factors behind the data-generating process \cite{bengioRepresentationLearningReview2013}, rendering the definition of domains multifaceted. HMOE simply discovers an intuitive and digestible way of data partitioning in order to enhance its DG performance.

\subsection{Ablation Studies}

The role of the intra-domain \emph{mixup} has been validated before. In this section, we analyze the contribution of other components of HMOE through ablation studies, as shown in \cref{tab:ablation}. We use the silhouette coefficient (SC) to quantitatively evaluate the clustering of HMOE in terms of cluster compactness and separation. SC ranges from -1 (poor) to 1 (good). Clusters are identified by gate values and their distances are measured using the output of the D2V encoder.

\vspace{1em}
\begin{table}[htbp]
	\centering
	\small
	\setlength\tabcolsep{2pt}
	\begin{tabular}{ccccccccc}
		\hline
		Name & $\mathcal{L}_{en}$ & $\mathcal{L}_{kl}$ & $\mathcal{L}_{ad}$ & VLCS   & PACS   & Office   & TerraInc & Avg. SC \\
		\hline
		H1   & -                        &  -          &  -            & 78.0          & 86.8          & 68.4          & 50.5  &     0.37   \\
		H2   & -                        &  -          &  \checkmark   & 77.8          & \textbf{86.9} & 69.1          & 51.2  &     0.27   \\
		H3   & \checkmark  &  -          &  -                         & 77.3          & 84.8          & 69.0          & 48.2  &   Collapse \\
		H4   & \checkmark  &  -          &  \checkmark                & 77.8          & 86.3          & 68.6          & 49.2  &   Collapse \\
		H5   & \checkmark  &  \checkmark &  -                         & 77.7          & 86.8          & 68.7          & 50.5  &   0.65     \\
		H6   & \checkmark  &  \checkmark & \checkmark                 & \textbf{78.1} & 86.6          & \textbf{69.7} & \textbf{52.5}  &   0.60 \\
		\hline
	\end{tabular}
	\caption{Ablation studies for HMOE-ND (\checkmark means the corresponding loss is used and SC denotes the silhouette coefficient.)}
	\label{tab:ablation}
\end{table}

\noindent \textbf{Top-1 routing $\mathcal{L}_{en}$ and expert load balancing $\mathcal{L}_{kl}$} \\
\noindent The joint use of $\mathcal{L}_{en}$ and $\mathcal{L}_{kl}$ leads to better clustering with greater SC and promotes latent domain discovery. Without them, HMOE relies on the inherent soft partitioning of MoE. H6 outperforms H2 mostly, which could indicate that better clustering benefits DG performance. However, H1 and H5 perform similarly, probably due to the absence of $\mathcal{L}_{ad}$. We find that $\mathcal{L}_{en}$ without $\mathcal{L}_{kl}$ suffers from the learning collapse problem, \ie, some embedding vectors collapse together, leading to a drop in accuracy. An example is shown in \cref{fig:PACS_E8}. This demonstrates the importance of $\mathcal{L}_{kl}$.

\noindent \textbf{Class-adversarial training} \quad $\mathcal{L}_{ad}$ boosts accuracy in most cases, verifying the necessity of filtering out class-specific information from the D2V encoder. H2 and H6 have smaller SC than H1 and H5, respectively, which is reasonable since class information can still be used by H1 and H5 for clustering, but is somewhat diminished for H2 and H6 via $\mathcal{L}_{ad}$.

\subsection{More Empirical Analysis}
\begin{figure}[htbp]
    \centering
    \begin{subfigure}{0.32\linewidth}
        \includegraphics[width=\textwidth]{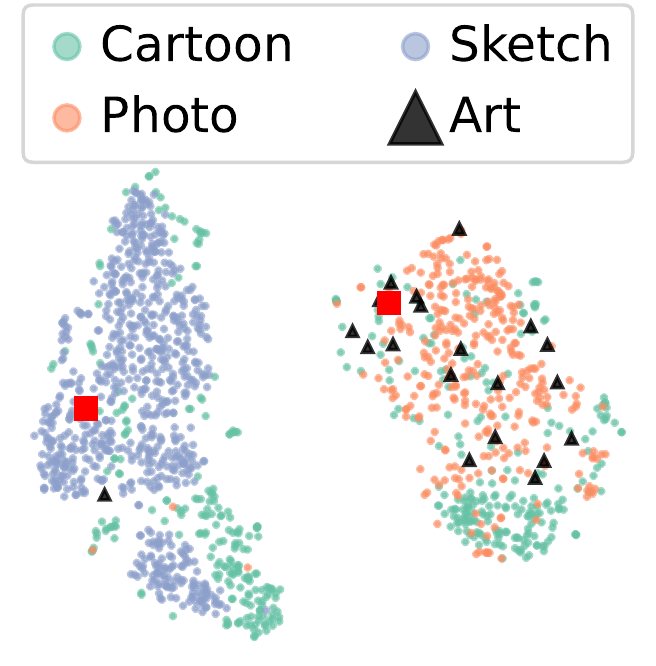}
        \caption{$K = 2$}
        \label{fig:PACS_E2}
    \end{subfigure}
    \begin{subfigure}{0.32\linewidth}
		\includegraphics[width=\textwidth]{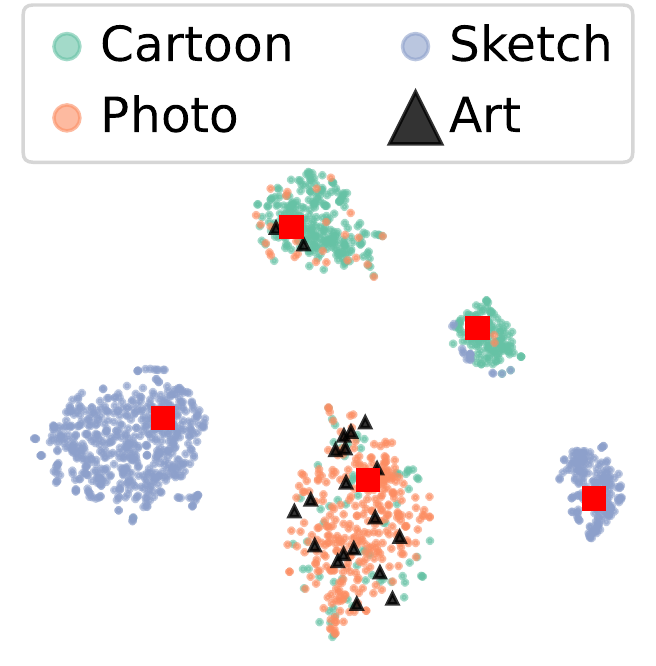}
		\caption{$K = 5$}
		\label{fig:PACS_E5}
	\end{subfigure}
    \begin{subfigure}{0.32\linewidth}
		\includegraphics[width=\textwidth]{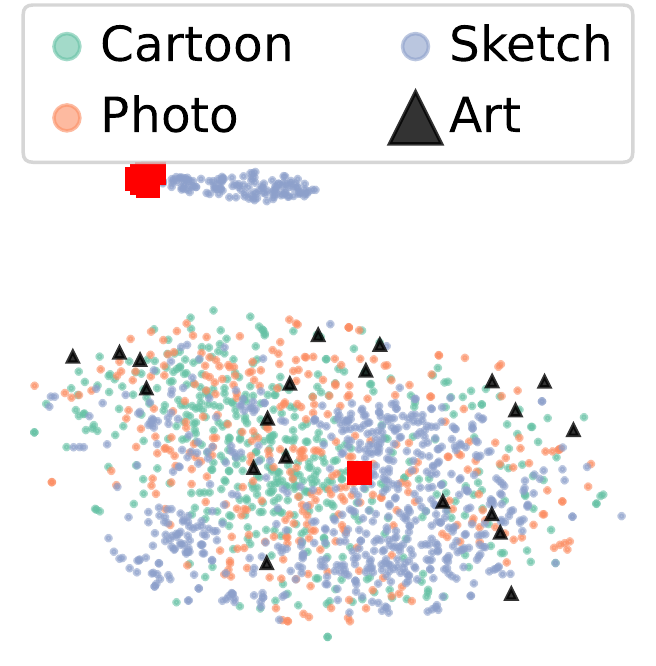}
		\caption{$K = 8$}
		\label{fig:PACS_E8}
	\end{subfigure}
    \caption{HMOE-ND using different $K$ for PACS (Art)}
    \label{fig:effect_K}
\end{figure}

\noindent \textbf{Effect of $K$ on latent domain discovery} \quad In \cref{fig:effect_K}, we try different numbers of embedding vectors $K$. For $K=2$, \emph{cartoon} is merged into \emph{sketch} and \emph{photo}. For $K=5$, \emph{sketch} and \emph{cartoon} are split into two sub-clusters. However, when $K$ increases to 8 and is much more than necessary, HMOE has difficulty in assigning data to different experts correctly and suffers from the learning collapse problem.

\noindent \textbf{Use Swin Transformer as featurizer} \quad \cite{liSparseFusionMixtureofExperts2022} investigated the impact of the backbone network (\ie, the featurizer for HMOE) on DG and found that transformer-based backbones outperform CNN-based counterparts. Motivated by this, we try Swin Transformer \cite{liuSwinTransformerHierarchical2021} (pretrained tiny version with similar complexity to ResNet-50 and its output size is 768) as featurizer (\cref{tab:swin}), which enhances both DeepAll and HMOE-MU, but the latter still performs much better.

\begin{table}[htbp]
	\centering
	\small
	\setlength\tabcolsep{3pt}
	\begin{tabular}{ccccc}
		\hline
                        & VLCS  & PACS & OfficeHome  & TerraInc \\
		\hline
		DeepAll     & 79.7  & 86.5 & 71.9        & 52.9     \\
	    HMOE-MU     & 79.8  & 88.1 & 74.6        & 54.7     \\
		\hline
	\end{tabular}
	\caption{Use Swin Transformer as featurizer of HMOE}
	\label{tab:swin}
\end{table}

\section{Conclusion}
\label{sec:conclusion}
This paper presents a novel DG method -- HMOE, which is based on Mixture of Experts, uses hypernetworks to generate the weights of experts, does not require domain labels, and enables latent domain discovery. HMOE achieves the SOTA performance in average accuracy on DomainBed.

However, it remains unclear how to effectively determine an appropriate number of experts or embedding vectors to fully explore domain information while avoiding the learning collapse. A promising solution that we will explore in future work is to use tree-structured hierarchical MoE to discover hierarchical domain knowledge, where each level contains only a number of experts but the number of multi-level inferred domains grows exponentially.

Finally, HMOE is versatile and scalable, and it should also be applicable to a wide range of problems beyond the scope of DG that are troubled by heterogeneous patterns.

\clearpage
\input{supplementary}

\clearpage
{
	\small
	\bibliographystyle{ieeenat_fullname}
	\bibliography{references}
}

\end{document}

%% file: preamble.tex
\usepackage[dvipsnames]{xcolor}


%% file: supplementary.tex
\setcounter{page}{1}

\appendix
\onecolumn
\begin{center}
	\Large{\textbf{HMOE: Hypernetwork-based Mixture of Experts for Domain Generalization}} \\
	\vspace{1.0em} \Large{Supplementary Material}\\
	\vspace{2.0em}
\end{center}

\section{Toy Regression Problem}
In the paper, we employ HMOE to address the domain generalization problem in image classification. In fact, HMOE is equally applicable to other problems troubled by heterogeneous patterns. To demonstrate the versatility of HMOE, we apply it to a toy regression task, aiming to learn a one-dimensional function defined over three intervals. Through this toy problem, we can also more intuitively understand the learning dynamics of HMOE, including the evolution of the gating mechanism and how experts become specialized gradually.

\begin{figure}[htbp]
	\centering
	\begin{subfigure}{0.6\linewidth}
		\includegraphics[width=\textwidth]{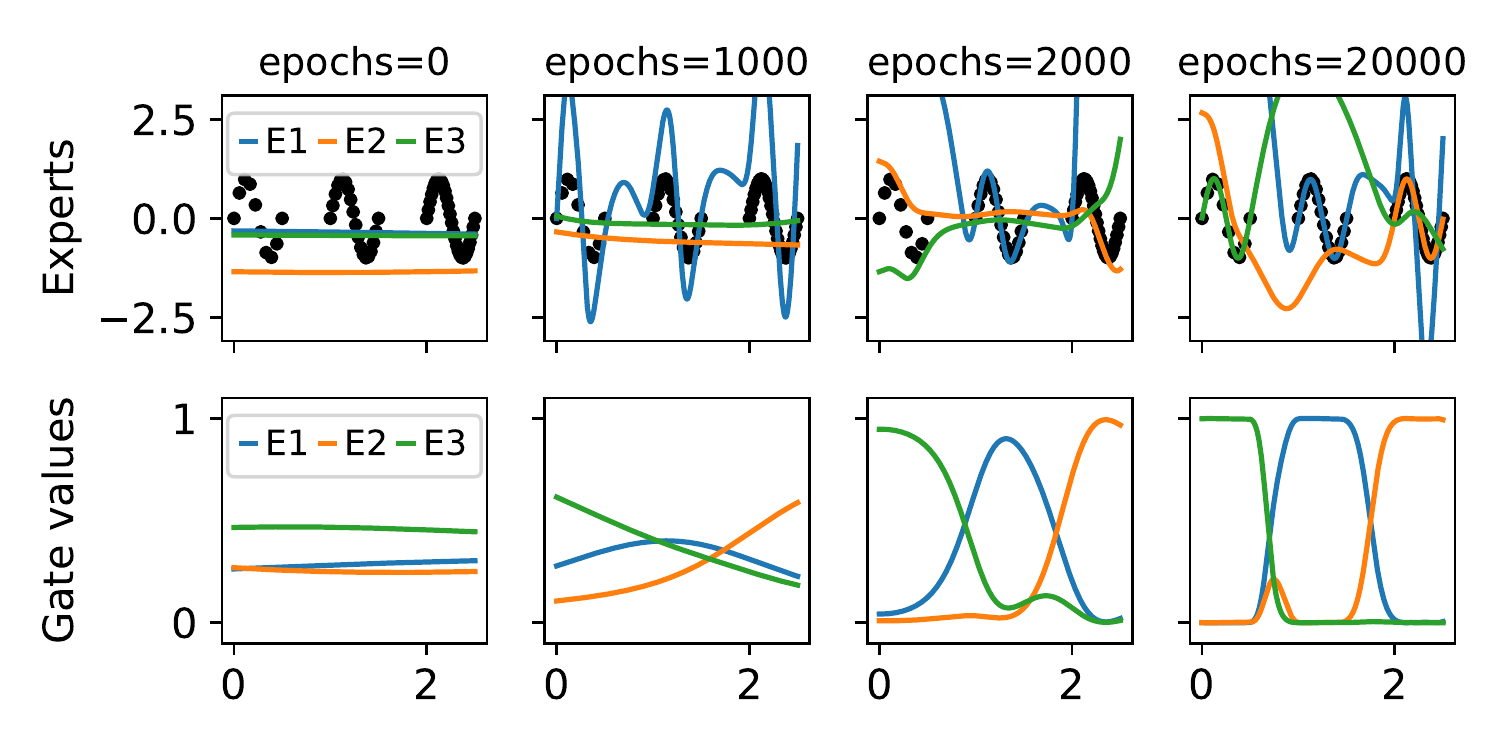}
		\caption{Experts' outputs and gate values during training}
		\label{fig:toy_experts}
	\end{subfigure}
	\begin{subfigure}{0.3\linewidth}
		\includegraphics[width=\textwidth]{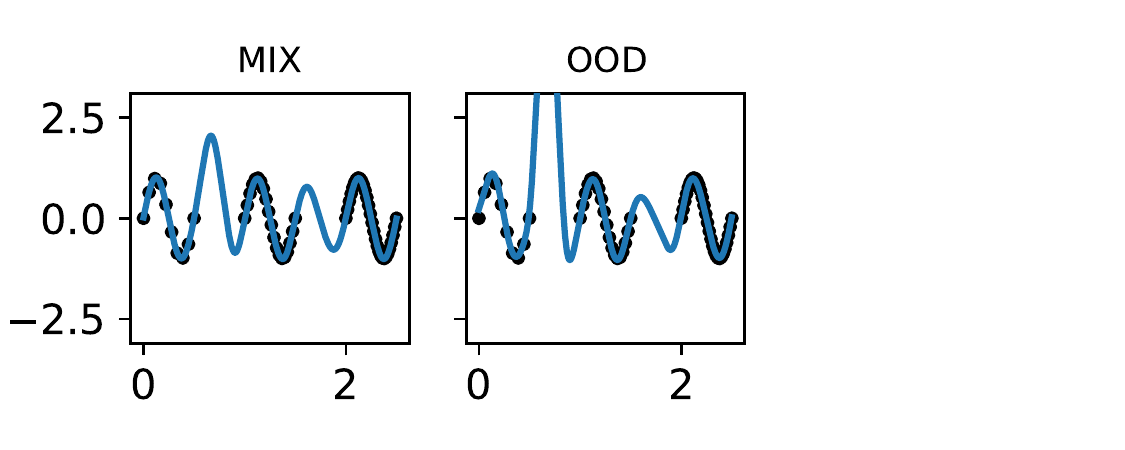}
		\caption{Two inference modes after training}
		\label{fig:toy_inference}
	\end{subfigure}
	\caption{A toy regression problem. We generate some data points using the function $y=\sin (4 \pi x)$ in three intervals and fit HMOE with three embedding vectors to these points. HMOE well identifies three intervals and experts also become specialized.}
	\label{fig:toy_regression}
\end{figure}

We use the function $y = \sin (4 \pi x)$ to generate 10, 20, and 30 data points uniformly in three intervals: $(0, 0.5)$, $(1, 1.5)$, and $(2, 2.5)$, respectively, as shown in \cref{fig:toy_experts}. Unequal data points are used to simulate a naturally unbalanced expert load. These three intervals represent three source domains, and we see if HMOE can generalize well in the regions between intervals.

HMOE uses three embedding vectors of dimension $D=8$, which are initialized using the standard normal distribution. All networks of HMOE are MLPs with 32 hidden units. The featurizer is a three-layer MLP whose input size is 1 and output size is 32. The encoder is a three-layer whose input size is 1 and output size is $D$. The classifier is a two-layer MLP whose input size is 32 and output size is 1. The hypernetwork is a four-layer MLP whose input size is $D$ and output size is the total number of learnable parameters (\ie, weights and biases) of the classifier. All MLPs use the SiLU activation function \cite{hendrycksGaussianErrorLinear2016} except the output layers. In addition, $\mathcal{L}_{y}$ (use MSE as the loss function), $\mathcal{L}_{en}$, and $\mathcal{L}_{kl}$ are used with $\lambda_{y} = \lambda_{en} = \lambda_{kl} = 1$, and HMOE is trained using Adam \cite{kingmaAdamMethodStochastic2014} with learning rate $0.001$ over $20,000$ epochs.

The evolution of the experts' outputs and gate values with respect to training epochs is depicted in \cref{fig:toy_experts}. From this, we can observe that three experts compete with each other and progressively delineate their respective positions. Notably, HMOE manages to identify three intervals even in the face of imbalanced data. After training, we compare two different inference modes, as shown in \cref{fig:toy_inference}. They all coincide well with the training points. MIX seems to perform better in the regions between intervals, while OOD presents an unexpected peak. Overall, HMOE demonstrates an ability to detect heterogeneous patterns within data.

\clearpage
\section{Description and visualization of datasets of DomainBed}

\begin{table}[htbp]
	\centering
	\small
	\setlength\tabcolsep{3pt}
	\begin{tabular}{cccccccccc}
		\toprule
		\textbf{Dataset} & \multicolumn{6}{c}{\textbf{Domains}}  & \textbf{\# of classes} & \textbf{\# of samples}   & \textbf{Image size}\\
		\toprule
		& \footnotesize{+90\%} & \footnotesize{+80\%} & \footnotesize{-90\%} & & & & & & \\
		Colored MNIST \cite{arjovskyInvariantRiskMinimization2019} &
		\raisebox{-.5\height}{\includegraphics[width=25pt, height=25pt]{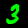}}   &
		\raisebox{-.5\height}{\includegraphics[width=25pt, height=25pt]{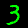}}    &
		\raisebox{-.5\height}{\includegraphics[width=25pt, height=25pt]{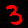}}   & & & & 2 & 70,000 & (2, 28, 28) \\
		& \multicolumn{6}{l}{\footnotesize{\emph{(degree of correlation between color and label)}}} & & & \\
		& \footnotesize{0$^{\circ}$} & \footnotesize{15$^{\circ}$} & \footnotesize{30$^{\circ}$} & \footnotesize{45$^{\circ}$} & \footnotesize{60$^{\circ}$} & \footnotesize{75$^{\circ}$} & & & \\
		Rotated MNIST \cite{ghifaryDomainGeneralizationObject2015} &
		\raisebox{-.5\height}{\includegraphics[width=25pt, height=25pt]{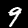}}   &
		\raisebox{-.5\height}{\includegraphics[width=25pt, height=25pt]{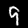}}   &
		\raisebox{-.5\height}{\includegraphics[width=25pt, height=25pt]{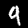}}  &
		\raisebox{-.5\height}{\includegraphics[width=25pt, height=25pt]{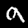}}   &
		\raisebox{-.5\height}{\includegraphics[width=25pt, height=25pt]{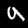}}   &
		\raisebox{-.5\height}{\includegraphics[width=25pt, height=25pt]{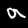}}  & 10 & 70,000 & (1, 28, 28) \\
		& \footnotesize{Caltech101} & \footnotesize{LabelMe} & \footnotesize{SUN09} & \footnotesize{VOC2007} & & & & & \\
		VLCS \cite{fangUnbiasedMetricLearning2013} &
		\raisebox{-.5\height}{\includegraphics[width=25pt, height=25pt]{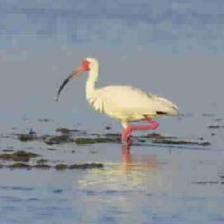}} &
		\raisebox{-.5\height}{\includegraphics[width=25pt, height=25pt]{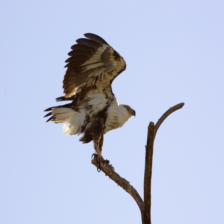}} &
		\raisebox{-.5\height}{\includegraphics[width=25pt, height=25pt]{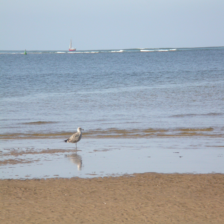}} &
		\raisebox{-.5\height}{\includegraphics[width=25pt, height=25pt]{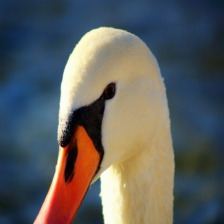}} & & & 5 & 10,729 & (3, 224, 224) \\
		& \footnotesize{Art} & \footnotesize{Cartoon} & \footnotesize{Photo} & \footnotesize{Sketch} & & & & & \\
		PACS \cite{liDeeperBroaderArtier2017} &
		\raisebox{-.5\height}{\includegraphics[width=25pt, height=25pt]{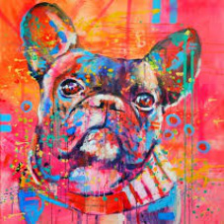}} &
		\raisebox{-.5\height}{\includegraphics[width=25pt, height=25pt]{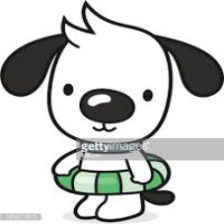}} &
		\raisebox{-.5\height}{\includegraphics[width=25pt, height=25pt]{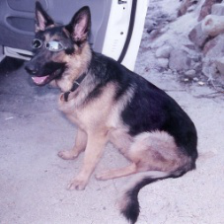}} &
		\raisebox{-.5\height}{\includegraphics[width=25pt, height=25pt]{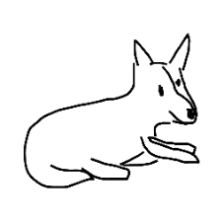}} & & & 7 & 9,991 & (3, 224, 224) \\
		& \footnotesize{Art} & \footnotesize{Clipart} & \footnotesize{Product} & \footnotesize{Photo} & & & & & \\
		OfficeHome \cite{venkateswaraDeepHashingNetwork2017} &
		\raisebox{-.5\height}{\includegraphics[width=25pt, height=25pt]{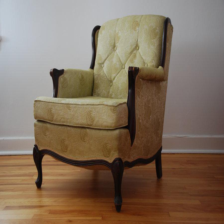}} &
		\raisebox{-.5\height}{\includegraphics[width=25pt, height=25pt]{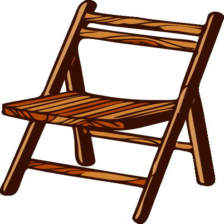}} &
		\raisebox{-.5\height}{\includegraphics[width=25pt, height=25pt]{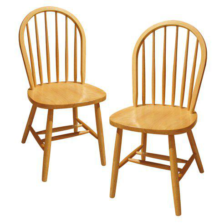}} &
		\raisebox{-.5\height}{\includegraphics[width=25pt, height=25pt]{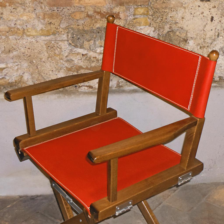}} & & & 65 & 15,588 & (3, 224, 224) \\
		& \footnotesize{L100} & \footnotesize{L38} & \footnotesize{L43} & \footnotesize{L46} & & & & & \\
		TerraIncognita \cite{beeryRecognitionTerraIncognita2018} &
		\raisebox{-.5\height}{\includegraphics[width=25pt, height=25pt]{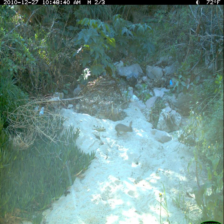}} &
		\raisebox{-.5\height}{\includegraphics[width=25pt, height=25pt]{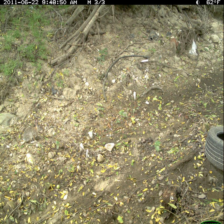}} &
		\raisebox{-.5\height}{\includegraphics[width=25pt, height=25pt]{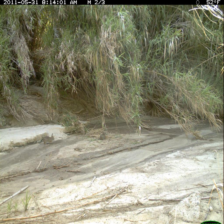}} &
		\raisebox{-.5\height}{\includegraphics[width=25pt, height=25pt]{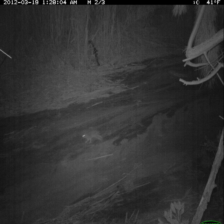}} & & & 10 & 24,788 & (3, 224, 224) \\
		& \multicolumn{6}{l}{\footnotesize{\emph{(camera trap location)}}} & & & \\
	\end{tabular}
	\label{table:datasets}
	\caption{Description and visualization of datasets used in our experiments (Adapted from \cite{gulrajaniSearchLostDomain2020})}
\end{table}

\section{Detailed domain generalization results}

We detail the domain generalization results for each dataset, and we format \textcolor{cyan}{first}, \underline{second} and \textcolor{gray}{worse than DeepAll} results.

\vspace{1em}
\begin{table}[htbp]
	\centering
	\small
	\setlength\tabcolsep{6pt}
	\begin{tabular}{lcccc|c}
		\hline
		\textbf{Algorithm} & \textbf{+90\%} & \textbf{+80\%} & \textbf{-90\%} & \textbf{Avg} & \textbf{Ranking} \\
		\hline
		\multicolumn{6}{c}{\emph{w/ Domain Labels}} \\
		\hline
		Mixup \cite{yanImproveUnsupervisedDomain2020} & 72.3 {\scriptsize$\pm$ 0.1} & 73.1 {\scriptsize$\pm$ 0.0} & 10.4 {\scriptsize$\pm$ 0.1} & 51.9 & 4 \\
		CORAL \cite{sunDeepCoralCorrelation2016} & \textcolor{gray}{71.3} {\scriptsize$\pm$ 0.3} & 73.0 {\scriptsize$\pm$ 0.2} & \textcolor{gray}{9.9} {\scriptsize$\pm$ 0.0} & 51.4 & 13 \\
		VREx \cite{kruegerOutofdistributionGeneralizationRisk2021} & \underline{73.1} {\scriptsize$\pm$ 0.3} & \underline{73.7} {\scriptsize$\pm$ 0.3} & 10.0 {\scriptsize$\pm$ 0.1} & \underline{52.2} & \underline{2} \\
		Fish \cite{shiGradientMatchingDomain2021a} & \textcolor{gray}{71.3} {\scriptsize$\pm$ 0.1} & 73.1 {\scriptsize$\pm$ 0.2} & 10.2 {\scriptsize$\pm$ 0.1} & 51.5 & 9 \\
		ARM \cite{zhangAdaptiveRiskMinimization2021} & \textcolor{cyan}{81.7} {\scriptsize$\pm$ 0.5} & \textcolor{cyan}{74.8} {\scriptsize$\pm$ 1.1} & 10.3 {\scriptsize$\pm$ 0.2} & \textcolor{cyan}{55.6} & \textcolor{cyan}{1} \\
		MTL \cite{blanchardDomainGeneralizationMarginal2021} & 71.6 {\scriptsize$\pm$ 0.3} & 72.9 {\scriptsize$\pm$ 0.3} & 10.2 {\scriptsize$\pm$ 0.0} & 51.5 & 10 \\
		GroupDRO \cite{sagawaDistributionallyRobustNeural2020} & 73.0 {\scriptsize$\pm$ 0.1} & 73.0 {\scriptsize$\pm$ 0.4} & 10.2 {\scriptsize$\pm$ 0.3} & 52.1 & 3 \\
		MLDG \cite{liLearningGeneralizeMetalearning2018} & \textcolor{gray}{37.5} {\scriptsize$\pm$ 9.9} & \textcolor{gray}{56.4} {\scriptsize$\pm$ 5.2} & \textcolor{cyan}{38.8} {\scriptsize$\pm$ 8.1} & \textcolor{gray}{44.2} & \textcolor{gray}{16} \\
		MMD \cite{liDomainGeneralizationAdversarial2018} & \textcolor{gray}{53.9} {\scriptsize$\pm$ 2.7} & \textcolor{gray}{51.6} {\scriptsize$\pm$ 0.8} & 10.1 {\scriptsize$\pm$ 0.1} & \textcolor{gray}{38.5} & \textcolor{gray}{18} \\
		DANN \cite{ganinDomainadversarialTrainingNeural2016} & 72.5 {\scriptsize$\pm$ 0.1} & 72.7 {\scriptsize$\pm$ 0.2} & 10.1 {\scriptsize$\pm$ 0.1} & 51.8 & 5 \\
		IRM \cite{arjovskyInvariantRiskMinimization2019} & \textcolor{gray}{57.0} {\scriptsize$\pm$ 2.7} & \textcolor{gray}{57.2} {\scriptsize$\pm$ 4.9} & \textcolor{gray}{9.7} {\scriptsize$\pm$ 0.0} & \textcolor{gray}{41.3} & \textcolor{gray}{17} \\
		\cdashline{1-5}
		HMOE-DL  & \textcolor{gray}{71.5} {\scriptsize$\pm$ 0.4} & 72.9 {\scriptsize$\pm$ 0.1} & 10.2 {\scriptsize$\pm$ 0.0} & 51.5 & 11 \\
		\hline
		\multicolumn{6}{c}{\emph{w/o Domain Labels}} \\
		\hline
		SelfReg \cite{kimSelfregSelfsupervisedContrastive2021} & \textcolor{gray}{71.1} {\scriptsize$\pm$ 0.3} & 73.0 {\scriptsize$\pm$ 0.0} & 10.1 {\scriptsize$\pm$ 0.2} & 51.4 & 14 \\
		SagNet \cite{namReducingDomainGap2021} & 72.2 {\scriptsize$\pm$ 0.0} & 73.3 {\scriptsize$\pm$ 0.3} & 10.0 {\scriptsize$\pm$ 0.1} & 51.8 & 6 \\
		RSC \cite{huangSelfchallengingImprovesCrossdomain2020} & 72.1 {\scriptsize$\pm$ 0.3} & \textcolor{gray}{72.3} {\scriptsize$\pm$ 0.8} & 10.1 {\scriptsize$\pm$ 0.1} & 51.5 & 12 \\
		DeepAll \cite{vapnikNatureStatisticalLearning1999} & 71.6 {\scriptsize$\pm$ 0.1} & 72.7 {\scriptsize$\pm$ 0.2} & 10.0 {\scriptsize$\pm$ 0.1} & 51.4 & 15 \\
		\cdashline{1-5}
		HMOE-ND  & 71.8 {\scriptsize$\pm$ 0.1} & 73.0 {\scriptsize$\pm$ 0.1} & \underline{10.5} {\scriptsize$\pm$ 0.2} & 51.8 & 7 \\
		HMOE-MU  & 71.7 {\scriptsize$\pm$ 0.4} & 73.0 {\scriptsize$\pm$ 0.3} & 10.3 {\scriptsize$\pm$ 0.1} & 51.7 & 8 \\
		\hline
	\end{tabular}
	\caption{Domain generalization results on Colored MNIST}
\end{table}

\begin{table}[htbp]
	\centering
	\small
	\setlength\tabcolsep{6pt}
	\begin{tabular}{lccccccc|c}
		\hline
		\textbf{Algorithm} & \textbf{0} & \textbf{15} & \textbf{30} & \textbf{45} & \textbf{60} & \textbf{75} & \textbf{Avg} & \textbf{Ranking} \\
		\hline
		\multicolumn{9}{c}{\emph{w/ Domain Labels}} \\
		\hline
		Mixup \cite{yanImproveUnsupervisedDomain2020} & \textcolor{gray}{93.8} {\scriptsize$\pm$ 0.1} & 98.8 {\scriptsize$\pm$ 0.1} & 99.0 {\scriptsize$\pm$ 0.0} & 99.1 {\scriptsize$\pm$ 0.1} & 98.9 {\scriptsize$\pm$ 0.0} & \textcolor{gray}{95.9} {\scriptsize$\pm$ 0.2} & \textcolor{gray}{97.6} & \textcolor{gray}{13} \\
		CORAL \cite{sunDeepCoralCorrelation2016} & 95.8 {\scriptsize$\pm$ 0.2} & 98.5 {\scriptsize$\pm$ 0.1} & 99.1 {\scriptsize$\pm$ 0.0} & \textcolor{gray}{99.0} {\scriptsize$\pm$ 0.1} & \underline{99.1} {\scriptsize$\pm$ 0.0} & \underline{96.6} {\scriptsize$\pm$ 0.1} & 98.0 & 2 \\
		VREx \cite{kruegerOutofdistributionGeneralizationRisk2021} & 95.5 {\scriptsize$\pm$ 0.1} & \textcolor{gray}{98.3} {\scriptsize$\pm$ 0.2} & \textcolor{gray}{98.9} {\scriptsize$\pm$ 0.1} & \textcolor{gray}{98.9} {\scriptsize$\pm$ 0.0} & 98.9 {\scriptsize$\pm$ 0.0} & 96.4 {\scriptsize$\pm$ 0.1} & 97.8 & 7 \\
		Fish \cite{shiGradientMatchingDomain2021a} & 95.5 {\scriptsize$\pm$ 0.4} & 98.7 {\scriptsize$\pm$ 0.0} & 99.0 {\scriptsize$\pm$ 0.0} & 99.1 {\scriptsize$\pm$ 0.1} & 98.9 {\scriptsize$\pm$ 0.0} & 96.3 {\scriptsize$\pm$ 0.3} & 97.9 & 6 \\
		ARM \cite{zhangAdaptiveRiskMinimization2021} & \underline{95.9} {\scriptsize$\pm$ 0.1} & 98.8 {\scriptsize$\pm$ 0.0} & \textcolor{gray}{98.9} {\scriptsize$\pm$ 0.1} & 99.1 {\scriptsize$\pm$ 0.0} & 98.9 {\scriptsize$\pm$ 0.0} & 96.2 {\scriptsize$\pm$ 0.1} & \textcolor{cyan}{98.1} & \textcolor{cyan}{1} \\
		MTL \cite{blanchardDomainGeneralizationMarginal2021} & 95.2 {\scriptsize$\pm$ 0.2} & 98.6 {\scriptsize$\pm$ 0.1} & \underline{99.1} {\scriptsize$\pm$ 0.0} & \textcolor{gray}{98.9} {\scriptsize$\pm$ 0.1} & \textcolor{gray}{98.8} {\scriptsize$\pm$ 0.1} & \textcolor{gray}{96.1} {\scriptsize$\pm$ 0.1} & 97.8 & 8 \\
		GroupDRO \cite{sagawaDistributionallyRobustNeural2020} & \textcolor{gray}{94.9} {\scriptsize$\pm$ 0.2} & 98.6 {\scriptsize$\pm$ 0.1} & \textcolor{gray}{98.9} {\scriptsize$\pm$ 0.0} & \textcolor{gray}{99.0} {\scriptsize$\pm$ 0.1} & 99.0 {\scriptsize$\pm$ 0.0} & 96.3 {\scriptsize$\pm$ 0.1} & 97.8 & 9 \\
		MLDG \cite{liLearningGeneralizeMetalearning2018} & 95.3 {\scriptsize$\pm$ 0.1} & 98.5 {\scriptsize$\pm$ 0.1} & 99.0 {\scriptsize$\pm$ 0.0} & \textcolor{gray}{99.0} {\scriptsize$\pm$ 0.0} & 98.9 {\scriptsize$\pm$ 0.1} & \textcolor{gray}{96.1} {\scriptsize$\pm$ 0.1} & 97.8 & 10 \\
		MMD \cite{liDomainGeneralizationAdversarial2018} & 95.8 {\scriptsize$\pm$ 0.3} & 98.8 {\scriptsize$\pm$ 0.0} & 99.0 {\scriptsize$\pm$ 0.1} & \textcolor{gray}{98.9} {\scriptsize$\pm$ 0.0} & 99.0 {\scriptsize$\pm$ 0.0} & 96.2 {\scriptsize$\pm$ 0.1} & 98.0 & 3 \\
		DANN \cite{ganinDomainadversarialTrainingNeural2016} & \textcolor{cyan}{95.9} {\scriptsize$\pm$ 0.1} & 98.5 {\scriptsize$\pm$ 0.1} & \textcolor{gray}{98.6} {\scriptsize$\pm$ 0.0} & \textcolor{gray}{98.8} {\scriptsize$\pm$ 0.0} & \textcolor{gray}{98.7} {\scriptsize$\pm$ 0.0} & \textcolor{gray}{95.6} {\scriptsize$\pm$ 0.1} & \textcolor{gray}{97.7} & \textcolor{gray}{12} \\
		IRM \cite{arjovskyInvariantRiskMinimization2019} & \textcolor{gray}{81.9} {\scriptsize$\pm$ 2.4} & \textcolor{gray}{88.1} {\scriptsize$\pm$ 4.2} & \textcolor{gray}{93.2} {\scriptsize$\pm$ 0.6} & \textcolor{gray}{91.3} {\scriptsize$\pm$ 2.8} & \textcolor{gray}{93.1} {\scriptsize$\pm$ 0.7} & \textcolor{gray}{76.0} {\scriptsize$\pm$ 0.7} & \textcolor{gray}{87.3} & \textcolor{gray}{18} \\
		\cdashline{1-8}
		HMOE-DL  & \textcolor{gray}{87.7} {\scriptsize$\pm$ 1.3} & \textcolor{gray}{93.3} {\scriptsize$\pm$ 2.2} & \textcolor{gray}{98.2} {\scriptsize$\pm$ 0.3} & \textcolor{gray}{98.6} {\scriptsize$\pm$ 0.0} & \textcolor{gray}{98.2} {\scriptsize$\pm$ 0.2} & \textcolor{gray}{88.8} {\scriptsize$\pm$ 1.5} & \textcolor{gray}{94.1} & \textcolor{gray}{17} \\
		\hline
		\multicolumn{9}{c}{\emph{w/o Domain Labels}} \\
		\hline
		SelfReg \cite{kimSelfregSelfsupervisedContrastive2021} & 95.7 {\scriptsize$\pm$ 0.1} & 98.7 {\scriptsize$\pm$ 0.0} & 99.0 {\scriptsize$\pm$ 0.0} & \textcolor{cyan}{99.2} {\scriptsize$\pm$ 0.0} & \textcolor{cyan}{99.1} {\scriptsize$\pm$ 0.0} & 96.5 {\scriptsize$\pm$ 0.1} & 98.0 & 4 \\
		SagNet \cite{namReducingDomainGap2021} & 95.1 {\scriptsize$\pm$ 0.3} & \underline{98.8} {\scriptsize$\pm$ 0.0} & \textcolor{cyan}{99.1} {\scriptsize$\pm$ 0.0} & 99.1 {\scriptsize$\pm$ 0.1} & 99.0 {\scriptsize$\pm$ 0.0} & \textcolor{cyan}{96.7} {\scriptsize$\pm$ 0.1} & 98.0   & 5 \\
		RSC \cite{huangSelfchallengingImprovesCrossdomain2020} & \textcolor{gray}{94.0} {\scriptsize$\pm$ 0.3} & \textcolor{gray}{98.3} {\scriptsize$\pm$ 0.1} & 99.0 {\scriptsize$\pm$ 0.0} & \textcolor{gray}{98.9} {\scriptsize$\pm$ 0.0} & 98.9 {\scriptsize$\pm$ 0.0} & \textcolor{gray}{95.9} {\scriptsize$\pm$ 0.1} & \textcolor{gray}{97.5} & \textcolor{gray}{15} \\
		DeepAll \cite{vapnikNatureStatisticalLearning1999} & 95.0 {\scriptsize$\pm$ 0.4} & 98.5 {\scriptsize$\pm$ 0.2} & 99.0 {\scriptsize$\pm$ 0.0} & \underline{99.1} {\scriptsize$\pm$ 0.0} & 98.9 {\scriptsize$\pm$ 0.0} & 96.2 {\scriptsize$\pm$ 0.1} & 97.8 & 11 \\
		\cdashline{1-8}
		HMOE-ND  & \textcolor{gray}{94.5} {\scriptsize$\pm$ 0.1} & 98.5 {\scriptsize$\pm$ 0.1} & \textcolor{gray}{98.8} {\scriptsize$\pm$ 0.0} & \textcolor{gray}{98.7} {\scriptsize$\pm$ 0.0} & \textcolor{gray}{98.7} {\scriptsize$\pm$ 0.1} & \textcolor{gray}{95.7} {\scriptsize$\pm$ 0.3} & \textcolor{gray}{97.5} & \textcolor{gray}{16} \\
		HMOE-MU  & \textcolor{gray}{94.6} {\scriptsize$\pm$ 0.3} & \textcolor{cyan}{98.8} {\scriptsize$\pm$ 0.0} & \textcolor{gray}{98.9} {\scriptsize$\pm$ 0.1} & \textcolor{gray}{98.8} {\scriptsize$\pm$ 0.0} & \textcolor{gray}{98.8} {\scriptsize$\pm$ 0.1} & \textcolor{gray}{95.6} {\scriptsize$\pm$ 0.2} & \textcolor{gray}{97.6} & \textcolor{gray}{14} \\
		\hline
	\end{tabular}
	\caption{Domain generalization results on Rotated MNIST}
\end{table}

\begin{table}[htbp]
	\centering
	\small
	\setlength\tabcolsep{6pt}
	\begin{tabular}{lccccc|c}
		\hline
		\textbf{Algorithm}  &  \textbf{Caltech101}    & \textbf{LabelMe}     & \textbf{SUN09}     & \textbf{VOC2007}  & \textbf{Avg} & \textbf{Ranking} \\
		\hline
		\multicolumn{7}{c}{\emph{w/ Domain Labels}} \\
		\hline
		Mixup \cite{yanImproveUnsupervisedDomain2020} & \textcolor{cyan}{98.2} {\scriptsize$\pm$ 0.3} & \textcolor{gray}{64.8} {\scriptsize$\pm$ 0.3} & 74.9 {\scriptsize$\pm$ 0.2} & \textcolor{gray}{76.9} {\scriptsize$\pm$ 1.0} & 78.7 & 3 \\
		CORAL \cite{sunDeepCoralCorrelation2016} & 97.2 {\scriptsize$\pm$ 0.4} & 65.8 {\scriptsize$\pm$ 0.4} & 74.0 {\scriptsize$\pm$ 0.3} & \textcolor{gray}{75.4} {\scriptsize$\pm$ 0.8} & 78.1 & 6 \\
		VREx \cite{kruegerOutofdistributionGeneralizationRisk2021} & 96.1 {\scriptsize$\pm$ 0.5} & \textcolor{gray}{64.8} {\scriptsize$\pm$ 1.2} & 72.6 {\scriptsize$\pm$ 0.5} & \textcolor{gray}{75.5} {\scriptsize$\pm$ 1.0} & \textcolor{gray}{77.3} & \textcolor{gray}{14} \\
		Fish \cite{shiGradientMatchingDomain2021a} & 96.8 {\scriptsize$\pm$ 0.5} & \textcolor{gray}{64.5} {\scriptsize$\pm$ 0.3} & 74.9 {\scriptsize$\pm$ 0.3} & \textcolor{gray}{76.1} {\scriptsize$\pm$ 1.0} & 78.1 & 7 \\
		ARM \cite{zhangAdaptiveRiskMinimization2021} & 97.0 {\scriptsize$\pm$ 0.2} & \underline{65.9} {\scriptsize$\pm$ 1.4} & 73.0 {\scriptsize$\pm$ 0.1} & \textcolor{gray}{76.2} {\scriptsize$\pm$ 1.4} & 78.0 & 9 \\
		MTL \cite{blanchardDomainGeneralizationMarginal2021} & 96.3 {\scriptsize$\pm$ 0.1} & \textcolor{gray}{64.5} {\scriptsize$\pm$ 0.3} & 72.6 {\scriptsize$\pm$ 0.5} & \textcolor{gray}{75.6} {\scriptsize$\pm$ 0.9} & \textcolor{gray}{77.3} & \textcolor{gray}{15} \\
		GroupDRO \cite{sagawaDistributionallyRobustNeural2020} & 97.1 {\scriptsize$\pm$ 0.3} & \textcolor{cyan}{65.9} {\scriptsize$\pm$ 0.7} & 72.4 {\scriptsize$\pm$ 1.7} & \textcolor{gray}{75.8} {\scriptsize$\pm$ 0.4} & 77.8 & 10 \\
		MLDG \cite{liLearningGeneralizeMetalearning2018} & 96.9 {\scriptsize$\pm$ 0.6} & \textcolor{gray}{61.5} {\scriptsize$\pm$ 0.8} & \textcolor{gray}{71.7} {\scriptsize$\pm$ 0.7} & \textcolor{gray}{76.5} {\scriptsize$\pm$ 0.2} & \textcolor{gray}{76.6} & \textcolor{gray}{17} \\
		MMD \cite{liDomainGeneralizationAdversarial2018} & 96.9 {\scriptsize$\pm$ 0.5} & \textcolor{gray}{64.2} {\scriptsize$\pm$ 1.9} & \textcolor{gray}{71.7} {\scriptsize$\pm$ 0.9} & \textcolor{gray}{76.6} {\scriptsize$\pm$ 1.8} & \textcolor{gray}{77.4} & \textcolor{gray}{13} \\
		DANN \cite{ganinDomainadversarialTrainingNeural2016} & 95.8 {\scriptsize$\pm$ 1.0} & \textcolor{gray}{65.1} {\scriptsize$\pm$ 0.7} & \textcolor{gray}{68.1} {\scriptsize$\pm$ 2.4} & \textcolor{gray}{73.5} {\scriptsize$\pm$ 0.7} & \textcolor{gray}{75.6} & \textcolor{gray}{18} \\
		IRM \cite{arjovskyInvariantRiskMinimization2019} & 96.8 {\scriptsize$\pm$ 0.3} & \textcolor{gray}{64.6} {\scriptsize$\pm$ 1.2} & 75.2 {\scriptsize$\pm$ 0.8} & \textcolor{gray}{76.6} {\scriptsize$\pm$ 3.4} & 78.3 & 5 \\
		\cdashline{1-6}
		HMOE-DL  & 95.5 {\scriptsize$\pm$ 1.4} & \textcolor{gray}{63.5} {\scriptsize$\pm$ 0.5} & 73.8 {\scriptsize$\pm$ 1.0} & \textcolor{gray}{75.0} {\scriptsize$\pm$ 1.5} & \textcolor{gray}{77.0} & \textcolor{gray}{16} \\
		\hline
		\multicolumn{7}{c}{\emph{w/o Domain Labels}} \\
		\hline
		SelfReg \cite{kimSelfregSelfsupervisedContrastive2021} & \underline{97.6} {\scriptsize$\pm$ 0.4} & \textcolor{gray}{65.2} {\scriptsize$\pm$ 0.2} & \underline{75.5} {\scriptsize$\pm$ 0.2} & \textcolor{gray}{77.1} {\scriptsize$\pm$ 0.7} & \textcolor{cyan}{78.9} & \textcolor{cyan}{1} \\
		SagNet \cite{namReducingDomainGap2021} & 96.8 {\scriptsize$\pm$ 0.1} & \textcolor{gray}{63.0} {\scriptsize$\pm$ 1.0} & 72.3 {\scriptsize$\pm$ 0.2} & \textcolor{cyan}{78.7} {\scriptsize$\pm$ 1.1} & 77.7 & 11 \\
		RSC \cite{huangSelfchallengingImprovesCrossdomain2020} & 96.7 {\scriptsize$\pm$ 0.9} & \textcolor{gray}{64.7} {\scriptsize$\pm$ 0.7} & \textcolor{cyan}{76.4} {\scriptsize$\pm$ 0.6} & \textcolor{gray}{77.4} {\scriptsize$\pm$ 0.8} & \underline{78.8} & \underline{2} \\
		DeepAll \cite{vapnikNatureStatisticalLearning1999} & 95.0 {\scriptsize$\pm$ 0.5} & 65.4 {\scriptsize$\pm$ 1.0} & 72.0 {\scriptsize$\pm$ 1.2} & 77.7 {\scriptsize$\pm$ 0.3} & 77.5 & 12 \\
		\cdashline{1-6}
		HMOE-ND  & 96.8 {\scriptsize$\pm$ 0.5} & \textcolor{gray}{64.7} {\scriptsize$\pm$ 0.5} & 75.0 {\scriptsize$\pm$ 0.1} & \textcolor{gray}{76.1} {\scriptsize$\pm$ 1.5} & 78.1 & 8 \\
		HMOE-MU  & 97.1 {\scriptsize$\pm$ 0.2} & \textcolor{gray}{64.6} {\scriptsize$\pm$ 0.7} & 74.9 {\scriptsize$\pm$ 0.4} & \underline{77.9} {\scriptsize$\pm$ 0.3} & 78.6 & 4 \\
		\hline
	\end{tabular}
	\caption{Domain generalization results on VLCS}
\end{table}

\begin{table}[htbp]
	\centering
	\small
	\setlength\tabcolsep{6pt}
	\begin{tabular}{lccccc|c}
		\hline
		\textbf{Algorithm}  &  \textbf{Art}    & \textbf{Cartoon}    & \textbf{Photo}    & \textbf{Sketch} & \textbf{Avg} & \textbf{Ranking} \\
		\hline
		\multicolumn{7}{c}{\emph{w/ Domain Labels}} \\
		\hline
		Mixup \cite{yanImproveUnsupervisedDomain2020} & 88.1 {\scriptsize$\pm$ 0.3} & 81.7 {\scriptsize$\pm$ 1.0} & 98.1 {\scriptsize$\pm$ 0.1} & 78.6 {\scriptsize$\pm$ 1.6} & 86.6 & 6 \\
		CORAL \cite{sunDeepCoralCorrelation2016} & 87.8 {\scriptsize$\pm$ 0.9} & \underline{82.7} {\scriptsize$\pm$ 0.9} & 98.0 {\scriptsize$\pm$ 0.1} & 78.4 {\scriptsize$\pm$ 1.8} & 86.7 & 5 \\
		VREx \cite{kruegerOutofdistributionGeneralizationRisk2021} & 86.5 {\scriptsize$\pm$ 2.0} & \textcolor{gray}{79.2} {\scriptsize$\pm$ 0.9} & 97.7 {\scriptsize$\pm$ 0.3} & 80.6 {\scriptsize$\pm$ 1.2} & 86.0 & 10 \\
		Fish \cite{shiGradientMatchingDomain2021a} & \textcolor{gray}{86.0} {\scriptsize$\pm$ 1.8} & \textcolor{cyan}{83.1} {\scriptsize$\pm$ 0.3} & 98.1 {\scriptsize$\pm$ 0.3} & 80.5 {\scriptsize$\pm$ 2.3} & 86.9 & 4 \\
		ARM \cite{zhangAdaptiveRiskMinimization2021} & \textcolor{gray}{86.2} {\scriptsize$\pm$ 1.2} & \textcolor{gray}{81.5} {\scriptsize$\pm$ 0.7} & \textcolor{gray}{97.2} {\scriptsize$\pm$ 0.3} & 77.9 {\scriptsize$\pm$ 1.1} & \textcolor{gray}{85.7} & \textcolor{gray}{12} \\
		MTL \cite{blanchardDomainGeneralizationMarginal2021} & 88.4 {\scriptsize$\pm$ 0.8} & \textcolor{gray}{80.7} {\scriptsize$\pm$ 1.2} & 97.8 {\scriptsize$\pm$ 0.2} & \textcolor{gray}{75.2} {\scriptsize$\pm$ 1.8} & \textcolor{gray}{85.5} & \textcolor{gray}{13} \\
		GroupDRO \cite{sagawaDistributionallyRobustNeural2020} & \textcolor{gray}{86.3} {\scriptsize$\pm$ 1.9} & \textcolor{gray}{81.0} {\scriptsize$\pm$ 0.6} & 97.8 {\scriptsize$\pm$ 0.1} & \textcolor{gray}{74.9} {\scriptsize$\pm$ 2.0} & \textcolor{gray}{85.0} & \textcolor{gray}{15} \\
		MLDG \cite{liLearningGeneralizeMetalearning2018} & \textcolor{cyan}{90.7} {\scriptsize$\pm$ 0.3} & \textcolor{gray}{80.4} {\scriptsize$\pm$ 0.4} & 97.9 {\scriptsize$\pm$ 0.1} & 79.5 {\scriptsize$\pm$ 0.8} & \underline{87.1} & \underline{2} \\
		MMD \cite{liDomainGeneralizationAdversarial2018} & 87.0 {\scriptsize$\pm$ 0.4} & \textcolor{gray}{79.6} {\scriptsize$\pm$ 0.9} & \textcolor{gray}{97.4} {\scriptsize$\pm$ 0.3} & \textcolor{gray}{72.6} {\scriptsize$\pm$ 1.8} & \textcolor{gray}{84.2} & \textcolor{gray}{16} \\
		DANN \cite{ganinDomainadversarialTrainingNeural2016} & \textcolor{gray}{79.4} {\scriptsize$\pm$ 1.9} & \textcolor{gray}{74.7} {\scriptsize$\pm$ 0.9} & \textcolor{gray}{97.0} {\scriptsize$\pm$ 1.1} & \textcolor{gray}{57.1} {\scriptsize$\pm$ 7.0} & \textcolor{gray}{77.0} & \textcolor{gray}{18} \\
		IRM \cite{arjovskyInvariantRiskMinimization2019} & \textcolor{gray}{84.8} {\scriptsize$\pm$ 1.8} & \textcolor{gray}{73.9} {\scriptsize$\pm$ 1.9} & \textcolor{cyan}{98.6} {\scriptsize$\pm$ 0.1} & \textcolor{gray}{71.3} {\scriptsize$\pm$ 1.0} & \textcolor{gray}{82.1} & \textcolor{gray}{17} \\
		\cdashline{1-6}
		HMOE-DL  & 87.5 {\scriptsize$\pm$ 1.4} & \textcolor{gray}{78.9} {\scriptsize$\pm$ 1.3} & 97.6 {\scriptsize$\pm$ 0.1} & 77.9 {\scriptsize$\pm$ 1.3} & \textcolor{gray}{85.5} & \textcolor{gray}{14} \\
		\hline
		\multicolumn{7}{c}{\emph{w/o Domain Labels}} \\
		\hline
		SelfReg \cite{kimSelfregSelfsupervisedContrastive2021} & 86.8 {\scriptsize$\pm$ 2.0} & 82.3 {\scriptsize$\pm$ 0.8} & 97.6 {\scriptsize$\pm$ 0.2} & 77.8 {\scriptsize$\pm$ 0.8} & 86.1 & 9 \\
		SagNet \cite{namReducingDomainGap2021} & \textcolor{gray}{85.3} {\scriptsize$\pm$ 2.0} & 81.8 {\scriptsize$\pm$ 1.6} & 97.7 {\scriptsize$\pm$ 0.3} & 79.8 {\scriptsize$\pm$ 0.8} & 86.2 & 8 \\
		RSC \cite{huangSelfchallengingImprovesCrossdomain2020} & 86.6 {\scriptsize$\pm$ 1.2} & 82.4 {\scriptsize$\pm$ 0.4} & \textcolor{gray}{97.4} {\scriptsize$\pm$ 0.3} & \underline{81.6} {\scriptsize$\pm$ 0.7} & 87.0 & 3 \\
		DeepAll \cite{vapnikNatureStatisticalLearning1999} & 86.4 {\scriptsize$\pm$ 1.2} & 81.7 {\scriptsize$\pm$ 0.6} & 97.5 {\scriptsize$\pm$ 0.3} & 77.7 {\scriptsize$\pm$ 1.8} & 85.8 & 11 \\
		\cdashline{1-6}
		HMOE-ND  & 87.1 {\scriptsize$\pm$ 0.7} & 81.7 {\scriptsize$\pm$ 0.9} & 97.7 {\scriptsize$\pm$ 0.1} & 79.9 {\scriptsize$\pm$ 1.1} & 86.6 & 7 \\
		HMOE-MU  & \underline{89.6} {\scriptsize$\pm$ 0.5} & \textcolor{gray}{81.2} {\scriptsize$\pm$ 1.0} & \underline{98.3} {\scriptsize$\pm$ 0.2} & \textcolor{cyan}{82.9} {\scriptsize$\pm$ 1.6} & \textcolor{cyan}{88.0} & \textcolor{cyan}{1} \\
		\hline
	\end{tabular}
	\caption{Domain generalization results on PACS}
\end{table}

\begin{table}[htbp]
\centering
\small
\setlength\tabcolsep{6pt}
\begin{tabular}{lccccc|c}
\hline
\textbf{Algorithm}  & \textbf{Art}     & \textbf{Clipart}     & \textbf{Product}       & \textbf{Real}     & \textbf{Avg} & \textbf{Ranking} \\
\hline
\multicolumn{7}{c}{\emph{w/ Domain Labels}} \\
\hline
Mixup \cite{yanImproveUnsupervisedDomain2020} & 68.1 {\scriptsize$\pm$ 0.8} & 55.9 {\scriptsize$\pm$ 0.8} & 80.3 {\scriptsize$\pm$ 0.1} & \underline{82.0} {\scriptsize$\pm$ 0.3} & 71.6 & 3 \\
CORAL \cite{sunDeepCoralCorrelation2016} & \textcolor{cyan}{69.9} {\scriptsize$\pm$ 0.7} & \underline{56.8} {\scriptsize$\pm$ 0.1} & \underline{80.5} {\scriptsize$\pm$ 0.4} & 81.7 {\scriptsize$\pm$ 0.2} & \underline{72.2} & \underline{2} \\
VREx \cite{kruegerOutofdistributionGeneralizationRisk2021} & 66.4 {\scriptsize$\pm$ 0.8} & 54.0 {\scriptsize$\pm$ 0.4} & 78.2 {\scriptsize$\pm$ 0.2} & 80.6 {\scriptsize$\pm$ 0.2} & 69.8 & 5 \\
Fish \cite{shiGradientMatchingDomain2021a} & \textcolor{gray}{64.3} {\scriptsize$\pm$ 0.3} & 53.0 {\scriptsize$\pm$ 0.4} & 78.1 {\scriptsize$\pm$ 0.1} & \textcolor{gray}{79.4} {\scriptsize$\pm$ 0.7} & 68.7 & 10 \\
ARM \cite{zhangAdaptiveRiskMinimization2021} & \textcolor{gray}{60.4} {\scriptsize$\pm$ 0.2} & 52.2 {\scriptsize$\pm$ 0.6} & \textcolor{gray}{75.6} {\scriptsize$\pm$ 0.6} & \textcolor{gray}{77.9} {\scriptsize$\pm$ 0.3} & \textcolor{gray}{66.5} & \textcolor{gray}{15} \\
MTL \cite{blanchardDomainGeneralizationMarginal2021} & \textcolor{gray}{64.3} {\scriptsize$\pm$ 0.7} & \textcolor{gray}{52.1} {\scriptsize$\pm$ 1.3} & 78.5 {\scriptsize$\pm$ 0.1} & \textcolor{gray}{78.6} {\scriptsize$\pm$ 0.1} & \textcolor{gray}{68.4} & \textcolor{gray}{12} \\
GroupDRO \cite{sagawaDistributionallyRobustNeural2020} & \textcolor{gray}{63.7} {\scriptsize$\pm$ 0.8} & 52.9 {\scriptsize$\pm$ 0.8} & 77.6 {\scriptsize$\pm$ 0.2} & \textcolor{gray}{78.8} {\scriptsize$\pm$ 0.3} & \textcolor{gray}{68.3} & \textcolor{gray}{13} \\
MLDG \cite{liLearningGeneralizeMetalearning2018} & \textcolor{gray}{64.2} {\scriptsize$\pm$ 0.8} & 52.7 {\scriptsize$\pm$ 0.9} & 78.4 {\scriptsize$\pm$ 0.8} & \textcolor{gray}{78.1} {\scriptsize$\pm$ 0.2} & \textcolor{gray}{68.3} & \textcolor{gray}{14} \\
MMD \cite{liDomainGeneralizationAdversarial2018} & 65.6 {\scriptsize$\pm$ 0.3} & 53.7 {\scriptsize$\pm$ 0.5} & 77.8 {\scriptsize$\pm$ 0.1} & \textcolor{gray}{79.4} {\scriptsize$\pm$ 0.1} & 69.1 & 8 \\
DANN \cite{ganinDomainadversarialTrainingNeural2016} & \textcolor{gray}{62.0} {\scriptsize$\pm$ 0.9} & \textcolor{gray}{49.7} {\scriptsize$\pm$ 1.8} & \textcolor{gray}{76.1} {\scriptsize$\pm$ 0.5} & \textcolor{gray}{78.2} {\scriptsize$\pm$ 0.4} & \textcolor{gray}{66.5} & \textcolor{gray}{16} \\
IRM \cite{arjovskyInvariantRiskMinimization2019} & \textcolor{gray}{60.4} {\scriptsize$\pm$ 0.4} & \textcolor{gray}{49.6} {\scriptsize$\pm$ 1.0} & \textcolor{gray}{73.2} {\scriptsize$\pm$ 0.8} & \textcolor{gray}{76.2} {\scriptsize$\pm$ 0.5} & \textcolor{gray}{64.9} & \textcolor{gray}{18} \\
\cdashline{1-6}
HMOE-DL  & 64.8 {\scriptsize$\pm$ 0.7} & 53.0 {\scriptsize$\pm$ 1.4} & 78.6 {\scriptsize$\pm$ 0.3} & \textcolor{gray}{79.0} {\scriptsize$\pm$ 0.3} & 68.9 & 9 \\
\hline
\multicolumn{7}{c}{\emph{w/o Domain Labels}} \\
\hline
SelfReg \cite{kimSelfregSelfsupervisedContrastive2021} & 68.0 {\scriptsize$\pm$ 0.4} & 55.7 {\scriptsize$\pm$ 0.4} & 79.7 {\scriptsize$\pm$ 0.2} & 81.9 {\scriptsize$\pm$ 0.6} & 71.3 & 4 \\
SagNet \cite{namReducingDomainGap2021} & \textcolor{gray}{63.7} {\scriptsize$\pm$ 0.9} & 54.6 {\scriptsize$\pm$ 0.2} & 78.2 {\scriptsize$\pm$ 0.2} & 80.7 {\scriptsize$\pm$ 0.4} & 69.3 & 7 \\
RSC \cite{huangSelfchallengingImprovesCrossdomain2020} & \textcolor{gray}{60.7} {\scriptsize$\pm$ 1.4} & \textcolor{gray}{51.4} {\scriptsize$\pm$ 0.3} & \textcolor{gray}{74.8} {\scriptsize$\pm$ 1.1} & \textcolor{gray}{75.1} {\scriptsize$\pm$ 1.3} & \textcolor{gray}{65.5} & \textcolor{gray}{17} \\
DeepAll \cite{vapnikNatureStatisticalLearning1999} & 64.7 {\scriptsize$\pm$ 0.6} & 52.2 {\scriptsize$\pm$ 1.0} & 77.4 {\scriptsize$\pm$ 0.2} & 79.8 {\scriptsize$\pm$ 0.2} & 68.5 & 11 \\
\cdashline{1-6}
HMOE-ND  & 65.6 {\scriptsize$\pm$ 0.1} & 54.7 {\scriptsize$\pm$ 0.6} & 78.8 {\scriptsize$\pm$ 0.2} & 79.9 {\scriptsize$\pm$ 0.3} & 69.7 & 6 \\
HMOE-MU  & \underline{68.7} {\scriptsize$\pm$ 0.6} & \textcolor{cyan}{57.7} {\scriptsize$\pm$ 0.4} & \textcolor{cyan}{81.0} {\scriptsize$\pm$ 0.2} & \textcolor{cyan}{82.6} {\scriptsize$\pm$ 0.4} & \textcolor{cyan}{72.5} & \textcolor{cyan}{1} \\
\hline
\end{tabular}
\caption{Domain generalization results on OfficeHome}
\end{table}

\begin{table}[htbp]
	\centering
	\small
	\setlength\tabcolsep{6pt}
	\begin{tabular}{lccccc|c}
		\hline
		\textbf{Algorithm} & \textbf{L100} & \textbf{L38} & \textbf{L43} & \textbf{L46} & \textbf{Avg} & \textbf{Ranking} \\
		\hline
		\multicolumn{7}{c}{\emph{w/ Domain Labels}} \\
		\hline
		Mixup \cite{yanImproveUnsupervisedDomain2020} & \textcolor{cyan}{68.3} {\scriptsize$\pm$ 2.0} & 43.9 {\scriptsize$\pm$ 0.4} & \textcolor{gray}{56.9} {\scriptsize$\pm$ 1.5} & \textcolor{gray}{36.6} {\scriptsize$\pm$ 0.5} & 51.4 & 5 \\
		CORAL \cite{sunDeepCoralCorrelation2016} & 52.9 {\scriptsize$\pm$ 3.7} & 46.8 {\scriptsize$\pm$ 1.4} & 59.5 {\scriptsize$\pm$ 0.4} & \textcolor{gray}{36.3} {\scriptsize$\pm$ 0.9} & 48.9 & 15 \\
		VREx \cite{kruegerOutofdistributionGeneralizationRisk2021} & 60.7 {\scriptsize$\pm$ 1.7} & 44.8 {\scriptsize$\pm$ 1.2} & 58.9 {\scriptsize$\pm$ 1.4} & 42.6 {\scriptsize$\pm$ 1.3} & 51.8 & 3 \\
		Fish \cite{shiGradientMatchingDomain2021a} & 55.7 {\scriptsize$\pm$ 2.2} & 46.9 {\scriptsize$\pm$ 2.5} & \underline{59.9} {\scriptsize$\pm$ 0.4} & 41.3 {\scriptsize$\pm$ 2.1} & 51.0 & 7 \\
		ARM \cite{zhangAdaptiveRiskMinimization2021} & 56.0 {\scriptsize$\pm$ 3.1} & 44.3 {\scriptsize$\pm$ 1.4} & \textcolor{gray}{54.9} {\scriptsize$\pm$ 0.3} & \textcolor{gray}{38.6} {\scriptsize$\pm$ 0.6} & 48.5 & 16 \\
		MTL \cite{blanchardDomainGeneralizationMarginal2021} & 55.1 {\scriptsize$\pm$ 0.8} & \underline{51.3} {\scriptsize$\pm$ 2.3} & \textcolor{gray}{57.8} {\scriptsize$\pm$ 0.8} & 41.2 {\scriptsize$\pm$ 2.1} & 51.3 & 6 \\
		GroupDRO \cite{sagawaDistributionallyRobustNeural2020} & 51.9 {\scriptsize$\pm$ 2.9} & 45.4 {\scriptsize$\pm$ 1.8} & \textcolor{cyan}{60.8} {\scriptsize$\pm$ 0.7} & 40.2 {\scriptsize$\pm$ 0.3} & 49.6 & 12 \\
		MLDG \cite{liLearningGeneralizeMetalearning2018} & 57.6 {\scriptsize$\pm$ 3.3} & 46.2 {\scriptsize$\pm$ 1.2} & \textcolor{gray}{58.4} {\scriptsize$\pm$ 0.7} & \textcolor{gray}{37.5} {\scriptsize$\pm$ 0.8} & 49.9 & 11 \\
		MMD \cite{liDomainGeneralizationAdversarial2018} & 61.0 {\scriptsize$\pm$ 2.7} & 43.2 {\scriptsize$\pm$ 0.6} & \textcolor{gray}{57.5} {\scriptsize$\pm$ 1.5} & \textcolor{gray}{38.3} {\scriptsize$\pm$ 2.2} & 50.0 & 10 \\
		DANN \cite{ganinDomainadversarialTrainingNeural2016} & \textcolor{gray}{48.8} {\scriptsize$\pm$ 1.1} & \textcolor{gray}{38.1} {\scriptsize$\pm$ 3.9} & \textcolor{gray}{44.1} {\scriptsize$\pm$ 4.4} & \textcolor{gray}{38.9} {\scriptsize$\pm$ 2.4} & \textcolor{gray}{42.5} & \textcolor{gray}{18} \\
		IRM \cite{arjovskyInvariantRiskMinimization2019} & \textcolor{gray}{49.4} {\scriptsize$\pm$ 4.3} & 47.6 {\scriptsize$\pm$ 2.4} & \textcolor{gray}{58.4} {\scriptsize$\pm$ 1.6} & \textcolor{cyan}{47.8} {\scriptsize$\pm$ 1.5} & 50.8 & 8 \\
		\cdashline{1-6}
		HMOE-DL  & 56.1 {\scriptsize$\pm$ 1.9} & 48.1 {\scriptsize$\pm$ 1.2} & \textcolor{gray}{57.7} {\scriptsize$\pm$ 0.8} & \textcolor{gray}{36.5} {\scriptsize$\pm$ 1.3} & 49.6 & 13 \\
		\hline
		\multicolumn{7}{c}{\emph{w/o Domain Labels}} \\
		\hline
		SelfReg \cite{kimSelfregSelfsupervisedContrastive2021} & 59.0 {\scriptsize$\pm$ 2.4} & 46.0 {\scriptsize$\pm$ 1.1} & 59.6 {\scriptsize$\pm$ 1.7} & 41.5 {\scriptsize$\pm$ 1.1} & 51.5 & 4 \\
		SagNet \cite{namReducingDomainGap2021} & 59.6 {\scriptsize$\pm$ 1.3} & 46.3 {\scriptsize$\pm$ 1.1} & 59.8 {\scriptsize$\pm$ 0.7} & \textcolor{gray}{37.2} {\scriptsize$\pm$ 1.6} & 50.7 & 9 \\
		RSC \cite{huangSelfchallengingImprovesCrossdomain2020} & 51.7 {\scriptsize$\pm$ 6.4} & 46.4 {\scriptsize$\pm$ 0.7} & 59.1 {\scriptsize$\pm$ 0.9} & \textcolor{gray}{39.2} {\scriptsize$\pm$ 1.1} & 49.1 & 14 \\
		DeepAll \cite{vapnikNatureStatisticalLearning1999} & 50.0 {\scriptsize$\pm$ 3.4} & 42.3 {\scriptsize$\pm$ 1.6} & 58.5 {\scriptsize$\pm$ 1.0} & 39.9 {\scriptsize$\pm$ 2.3} & 47.7 & 17 \\
		\cdashline{1-6}
		HMOE-ND  & 60.7 {\scriptsize$\pm$ 3.6} & \textcolor{cyan}{53.2} {\scriptsize$\pm$ 1.5} & \textcolor{gray}{56.7} {\scriptsize$\pm$ 1.2} & \textcolor{gray}{39.6} {\scriptsize$\pm$ 0.3} & \underline{52.5} & \underline{2} \\
		HMOE-MU  & \underline{67.3} {\scriptsize$\pm$ 1.0} & 43.4 {\scriptsize$\pm$ 1.4} & \textcolor{gray}{57.4} {\scriptsize$\pm$ 0.7} & \underline{43.0} {\scriptsize$\pm$ 2.8} & \textcolor{cyan}{52.8} & \textcolor{cyan}{1} \\
		\hline
	\end{tabular}
	\caption{Domain generalization results on TerraInc}
\end{table}